%% file: main.tex
\documentclass[letterpaper]{article} 
\usepackage{aaai2027}  

\usepackage{times}
\usepackage{float}
\usepackage{multirow}
\usepackage{subcaption}
\usepackage{booktabs}
\usepackage[dvipsnames]{xcolor}
\usepackage{colortbl}

\usepackage[hyphens]{url}  
\usepackage{graphicx} 
\graphicspath{{images/}}

\usepackage{natbib}  
\usepackage{caption} 
\usepackage{algorithm}
\usepackage{algorithmic}
\usepackage{amsmath}

\usepackage{newfloat}
\usepackage{listings}
\DeclareCaptionStyle{ruled}{labelfont=normalfont,labelsep=colon,strut=off} 
\floatstyle{ruled}
\newfloat{listing}{tb}{lst}{}
\floatname{listing}{Listing}

\usepackage{booktabs}

\input{math_commands.tex}

\title{Feature Competition and Its Representational Origins in Vision–Language Models: Evidence from the Stroop Paradigm}

\author{
    Nurbüke Teker\textsuperscript{\rm 1},
    Rui Xiao\textsuperscript{\rm 1,\rm 2},
    Zeynep Akata\textsuperscript{\rm 1,\rm 2},
    Shuchen Wu\textsuperscript{\rm 3,\rm 4}
}
\affiliations{
    \textsuperscript{\rm 1}Technical University of Munich\\
    \textsuperscript{\rm 2}Munich Center for Machine Learning\\
    \textsuperscript{\rm 3}University of Washington\\
    \textsuperscript{\rm 4}Allen Institute\\
    shuchen.wu@alleninstitute.org
}

\begin{document}

\maketitle

\begin{abstract}
Vision–language models (VLMs) are often asked questions about images containing competing cues, where answering requires prioritizing one feature over another. We study how VLMs resolve such conflicts using the Stroop task—a classic paradigm in which a color word is printed in a conflicting ink color (e.g., the word RED printed in blue). We construct a Stroop benchmark and evaluate two contrastive encoders (CLIP and SigLIP-2) and six generative VLMs on the paradigm. Across architectures, VLMs show a persistent bias toward text: when word identity and ink color conflict, VLMs preferentially respond to the word, even when explicitly instructed to report the color until the “word” is no longer legible. We then trace this behavioral bias to internal representations. In CLIP’s embedding space, we observed that word identity is encoded along stronger and more separable directions than ink color. We then try to steer CLIP's behavior by manipulating the image embedding. We found that the representational imbalance of feature saliency predicts steerability: steering along word-related directions reliably shifts model preferences. Steering along color-related directions is far less effective. Our study suggests that features within the same image can exert systematically unequal influence on VLM behavior, and that this imbalance may root in feature embedding geometry. More broadly, our study suggests that feature saliency is therefore an important factor in understanding and auditing VLMs for applications where multiple visual features are jointly represented.\end{abstract}

\begin{links}
    \link{Code}{https://github.com/NurbukeTeker/vlm-stroop-conflict}
\end{links}


\begin{figure*}[t!]
    \centering
    \includegraphics[width=1.00\textwidth]{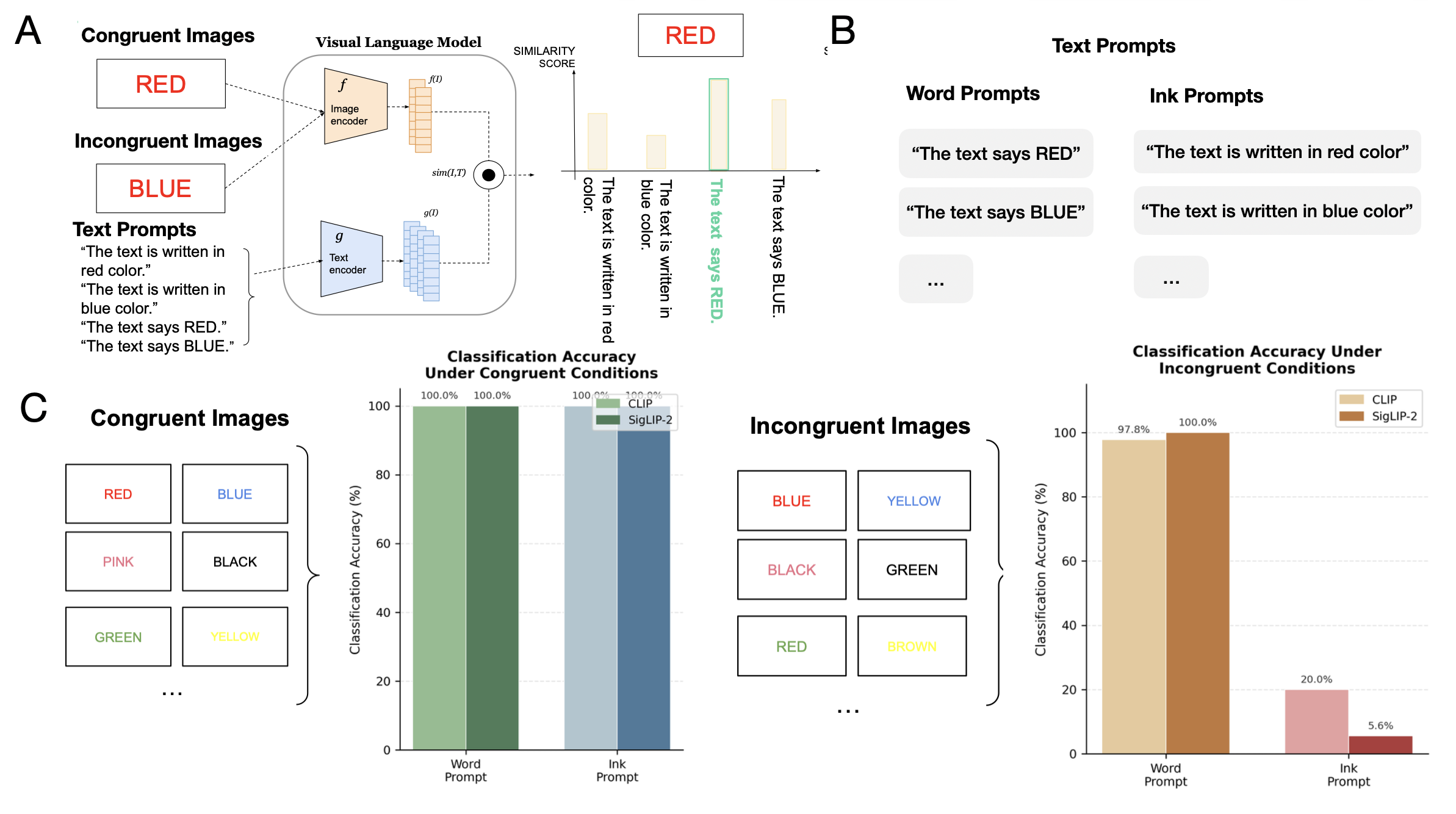}
\caption{\textbf{Studying feature competition in VLMs with the Stroop task.}
\textbf{(A)}~Example \emph{congruent} stimuli, in which word identity 
matches ink color, and \emph{incongruent} stimuli, in which the two 
conflict. An image encoder $f(I)$ and text encoder $g(T)$ map images 
and candidate prompts into a shared embedding space, where alignment 
is measured by cosine similarity $s(I,T)$.
\textbf{(B)}~Each image is scored against two prompt families: 
\emph{word-oriented} (e.g., ``The text says \textsc{blue}'') and 
\emph{ink-oriented} (e.g., ``The text is written in red'').
\textbf{(C)}~On congruent stimuli, both prompt families score 
correctly. On incongruent stimuli, word-oriented prompts still score 
correctly, but ink-oriented prompts do not --- the model aligns with 
the written word rather than the actual ink color.}
    \label{fig:clip_behavior}
\end{figure*}

\section{Introduction}
When visual and linguistic cues conflict, which one wins? A model reading a sign that says "safe" rendered in alarming red must somehow reconcile two opposing signals into a single response. Such conflicts matter for Vision–Language Models (VLMs)~\citep{Radford2021, Li2022BLIP, liu2023visualinstructiontuning}, which must resolve such conflicts implicitly, with no mechanism to explicitly weigh competing cues. Understanding which cues VLMs prioritize, and why, is therefore essential for predicting how they behave in real-world settings.

Psychology has long studied decision-making under conflicting cues. In the classical Stroop task~\citep{Stroop1935}, participants view color words (e.g., \texttt{RED}, \texttt{BLUE}) presented in incongruent ink colors and are instructed to report the ink color. When the word meaning and ink color conflicts, responses become slower and more error-prone. People's response tend to be strongly influenced by the text rather than ink color.

We adapt the Stroop paradigm to study how VLMs resolve conflicting visual cues. Our benchmark contains images in which a written word and its ink color either agree (\emph{congruent}) or conflict (\emph{incongruent}). We evaluate contrastive models, including CLIP~\citep{Radford2021}, SigLIP-2~\citep{tschannen2025siglip2multilingualvisionlanguage}, alongside open-source generative VLMs such as LLaVA. Across architectures, we find a consistent text-over-color bias: when word identity and ink color disagree, models are substantially more likely to respond according to the word.

To understand the source of this bias, we examine CLIP’s internal representation space and find that word-shape cues are encoded more strongly and more separably than color cues. This representational asymmetry is mirrored in steerability: interventions along word-related directions reliably shift model behavior, whereas color-related directions are weaker, more entangled, and substantially harder to manipulate. Together, these results suggest that behavioral cue dominance reflects the relative salience and separability of the corresponding internal representations. More broadly, our work demonstrates how classical psychological paradigms can be adapted to probe decision-making in multimodal foundation models, linking behavioral biases to the structure of their internal representations.

\section{Related Work}
\label{sec:related_work}
Vision-Language Models (VLMs) take images and texts as input and produce text-based outputs. They have become widely used across tasks such as image captioning~\citep{Xu2015, Hossain2019}, image–text retrieval~\citep{Faghri2018,xiao2025flair, kim2025cosmos}, visual question answering~\citep{Antol2015, Alayrac2022, Tsimpoukelli2021}, and others. 

However, a single image can support multiple, sometimes mutually exclusive, interpretations. For example, an image may contain a green stop sign: although its color typically signifies ``go,'' its text reads ``STOP.'' To act is to choose: when visual cues conflict, the perceptual system must resolve the evidence into a single task-relevant interpretation at a time~\citep{Kornmeier2012, Brascamp2021}. This problem has long been studied in human visual perception, particularly in object naming and categorization~\citep{smith2000, landau1988}. Although people can selectively attend to one feature, irrelevant variation in another feature can still interfere with their responses~\citep{Tanaka1999, navon1977353}.


Existing VLMs architectures fall into two broad paradigms. Embedding models, such as CLIP~\citep{Radford2021} and SigLIP~\citep{zhai2023sigmoid}, embed input images and texts into visual and textual embeddings,
aligning them in a shared embedding space. Despite alignment, these models do not support actual textual outputs. Meanwhile, generative models including BLIP-2~\citep{Li2022BLIP}, Flamingo~\citep{Alayrac2022}, LLaVA~\citep{liu2023visualinstructiontuning}, and Qwen2-VL~\citep{wang2024qwen2vlenhancingvisionlanguagemodels}, generate free-form text responses based on an input image and a text prompt.

While prior work has examined conflicts between input images and text 
prompts~\citep{li2023pope, deng2025wordsorvision, xiao2026finer,yuksekgonul2023when}, 
little is known about how VLMs resolve conflicting visual signals 
\emph{within} a single image. A related finding from vision models 
shows that when global shape and local texture conflict, CNNs classify 
by texture while humans rely on shape~\citep{geirhos2018imagenettrained} 
--- suggesting that cue biases are a general feature of learned visual models. 

Relatively little work has examined how internal representations give rise to biases toward the behavior of multi-modal models. The neuroscience literature provides several tools for studying representational structure, including Representational Dissimilarity analysis (RDMs) ~\citep{Kriegeskorte2008} that uses cosine-based similarity to study neural embedding space, and low-dimensional visualizations of embedding spaces. More recent interpretability literature methods extract interpretable ``chunks'' which encodes concepts from neural embeddings~\citep{wu2025discoveringchunksneuralembeddings} or identifies monosemantic features by using sparse autoencoders~\citep{Goh2021,Pach2025}. 

Prior work adapting the Stroop paradigm to VLMs \citep{zhao2025wcit} has focused on 
behavioral-level analysis, treating models as black boxes and 
stopping short of systematic image-level variations. Three questions 
remain open: whether the word-dominance effect persists under 
controlled stimulus manipulations, what representational properties 
give rise to it, and whether directly editing those representations 
can shift the model's feature preference.

\paragraph{Statement of Contributions}
This paper makes four contributions: (1) We introduce a multimodal Stroop benchmark that probes cue conflict across both contrastive and generative VLMs.(2) We systematically vary stimulus properties and identify a 
consistent text-over-color bias that persists across architectures 
and visual conditions.
(3)We trace this behavioral bias to representational asymmetries 
in CLIP's embedding space, where word identity is more salient than ink color.
(4) We show that these directions are steerable: editing them shifts 
the model's feature preference.
Our approach studying behavior and its representation origin allows us to explain not only \emph{that} word dominance occurs, but also \emph{why} it arises. 




\section{Evaluating VLMs on the Stroop Task}
\label{sec:stroop_dataset}
Following the classic Stroop paradigm~\citep{Stroop1935}, we started with constructing
a dataset of 100 images. Each image displays a color word (e.g., \textsc{red}) 
rendered in a specific ink color, which may or may not match the word's meaning. 
We use ten basic color categories: red, blue, green, yellow, orange, pink, 
purple, black, brown, and gray. We render each word in all ten ink colors, 
yielding 100 images in total: 10 \emph{congruent} (word matches ink color) 
and 90 \emph{incongruent} (word conflicts with ink color).

\subsection{Behavioral Analysis on Contrastive VLMs}
\label{sec:clip_siglip_behavioral}
For contrastive VLMs, we evaluate CLIP~\citep{Radford2021} and 
SigLIP-2~\citep{tschannen2025siglip2multilingualvisionlanguage}. Both models align image and text representations in a shared embedding space; 
CLIP uses a softmax-based contrastive objective, while SigLIP-2 replaces this 
with a sigmoid loss and additionally supports variable-resolution inputs.

Given an image $I$ and a set of candidate texts $\{T_k\}$, the model computes 
cosine similarity between the image embedding $f(I)$ and each text embedding 
$g(T_k)$:
\[
s(I, T_k) = \frac{f(I) \cdot g(T_k)}{\|f(I)\|\,\|g(T_k)\|}.
\]
The text prompt with the highest similarity score is taken as the model's prediction.

We evaluate each contrastive model using two prompt families (Figure~\ref{fig:clip_behavior}). \emph{Word-oriented} prompts ask about 
word identity: \textit{``The text says \textsc{red}''}; \emph{ink-oriented} 
prompts ask about ink color: \textit{``The text is written in \{ink\} color''}, 
where \texttt{\{ink\}} is instantiated with each of the ten candidate color 
terms (red, blue, \ldots, black). Each image is evaluated in two separate 
forward passes, one per prompt family. In each pass, the model receives all 
ten prompts of that family and selects the highest-scoring one as its prediction.

Using this criterion, we evaluated CLIP and SigLIP-2 on the Stroop images.
\emph{Word accuracy} is the proportion of images for which the top-scoring 
word-oriented prompt names the correct written word; \emph{ink accuracy} is 
the proportion for which the top-scoring ink-oriented prompt names the correct 
ink color. Because each family is scored independently, the two values need not 
sum to 100\%. 



\begin{figure*}[t!]
    \centering
\includegraphics[width=0.75\textwidth]{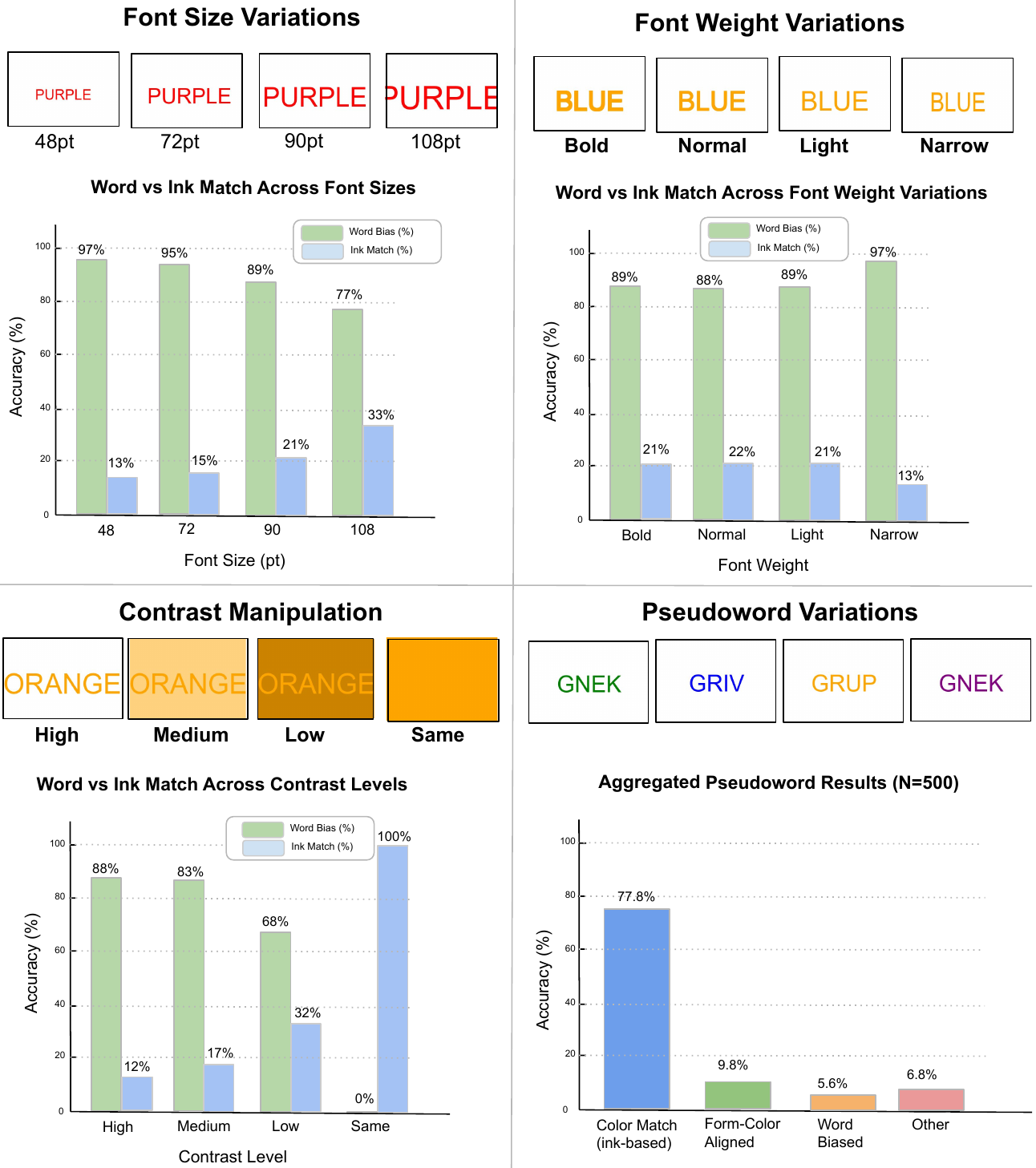}
    \caption{\textbf{Word bias persists under visual variations.}
    We test whether contrastive vlms's word bias persist varying different factors influencing conflicting visual features for incongruent stimuli: the font size (top left), the font weight (top right), the contrast (bottom left), and semantic meaning (bottom right).
    For every image, CLIP scores prompts of the form ``The word is written in \{ink\} color'', one per candidate color, and we take the highest-scoring prompt as its answer. We then check what that answer names: the actual ink color (\textbf{ink bias}), the color written in the image (\textbf{word bias}), or neither (\textbf{other}).
    The bottom row shows how often each outcome occurs in each condition, aggregated over all colors.
    }
    \label{fig:clip_visual_manipulations}
\end{figure*}
Shown in Figure \ref{fig:clip_behavior}, On congruent images, both CLIP and SigLIP-2 achieve ceiling performance for finding the correct word-prompt and color prompt. On incongruent images, the word-oriented evaluation reaches \textbf{97.8\%} accuracy with respect to the ground-truth word, while the ink-oriented evaluation dramatically drops to about \textbf{20\%} accuracy with respect to the ground-truth ink color (Figure~\ref{fig:clip_behavior}). For SigLIP-2, the word-oriented evaluation reaches \textbf{100\%} accuracy with respect to the ground-truth word, while the color query evaluation dramatically drops to about \textbf{5–6\%} accuracy.  I.e. when being asked about the word, CLIP and SigLIP-2 assigns higher score to the correct text describing the word. But when being asked about the color, CLIP and SigLIP-2 misassign the highest  matching score to a text describing the incorrect color. Under incongruent conditions, both contrastive models fail to assign the correct ink color when the text and color misalign.

\paragraph{Word Bias Persists With Visual Variations}
To test whether the word bias is specific to our initial stimuli, we evaluated CLIP under controlled visual manipulations of the Stroop images. We varied font size (48--108\, pt), font weight (light, normal, bold, narrow), contrast (high, medium, low, and a \emph{same} condition where letters blend with the background).

The word bias persists with the controlled image variations. CLIP accurately identifies the word(85--97\%) whenever it is legible but exerts strong tendency to misidentify the ink color(20--30\%) when there is a word displayed in the image. The performance gap narrows only when the word becomes unreadable (e.g., 108\,pt overflow, zero contrast), at which point the model can accurately identify the color. 

To control for lexical semantics, we also generated pseudoword stimuli: the color word was replaced by a scrambled-letter nonword, with font, 
size, position, ink color held fixed. This removes the word's semantic color meaning while preserving its low-level visual form.

For this variation, we classify the model's top prediction into response categories: \emph{ink-aligned} if it matches the true ink color, \emph{text-biased} if it follows the original word form or its associated color identity, and \emph{other} if it matches neither. Shown in Figure \ref{fig:clip_behavior}, under pseudoword conditions, CLIP shifts strongly toward ink-based predictions, with approximately 78\% of responses matching the ink color.

Beyond the controlled manipulations, we evaluated CLIP on a larger Stroop corpus of 23{,}338 images with systematically varied background texture, tone, saturation, brightness, and mild spatial perturbations, while preserving the word--ink conflict structure of the base dataset.
The observation from the smaller-scale Stroop dataset persistent: word accuracy stayed near-perfect (99.5\%), while ink recognition remained low (15.9\%). Additional generation details are provided in Appendix.

\subsection{Behavioral Analysis on Generative VLMs}
\label{sec:vlm_behavioral}
\begin{figure}[b!]
    \centering
    \includegraphics[width=\columnwidth]{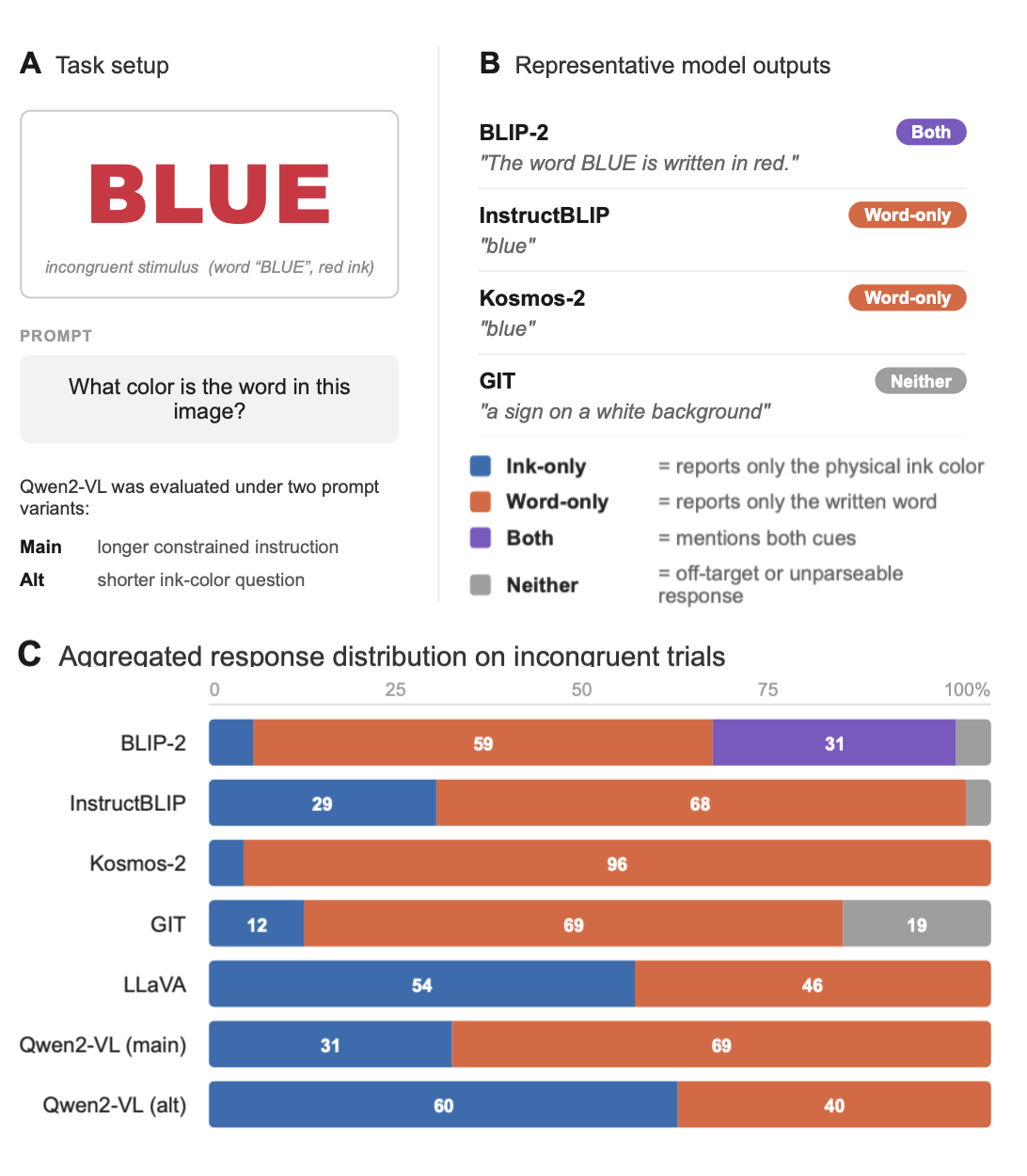}
    \caption{\textbf{Evaluating the Stroop Paradigm on Generative VLMs}
    (A) The task setup.
    We show each model an incongruent Stroop image and ask:
    \textit{``What color is the word in this image?''}
     For Qwen2-VL-7B-Instruct we also test two variants of this prompt, to check how much the phrasing matters.
    (B) Example answers from models and their categories. Each answer is categorized into one of four types: answer containing only the actual ink color (\textit{Ink-only}), only the word written (\textit{Word-only}), both (\textit{Both}), or neither (\textit{Neither}).
    (C) Distribution of answer types across all generative vlms.
    Most models mostly give \textit{Word-only} answers and they report the written word rather than the ink color. BLIP-2 often names both cues in one answer, LLaVA is the only model that reports the ink color more often than the word, and
    Qwen2-VL's answers change substantially with the prompt phrasing. Generative VLMs mostly answer with the written word, not the ink color.
    }
    \label{fig:generative_vlm_summary}
\end{figure}

We then evaluate generative Vision--Language Models on the Stroop task (Figure \ref{fig:generative_vlm_summary}). These models receive an image and an instruction, and generate free-form text responses. Specifically, we evaluate six open-source generative VLMs: BLIP-2 (FlanT5-XL)~\citep{Li2022BLIP}, InstructBLIP 
(Vicuna-7B)~\citep{dai2023instructblip}, Kosmos-2 
(1B)~\citep{peng2024grounding}, LLaVA 
(Vicuna-7B v1.6)~\citep{liu2023visualinstructiontuning}, 
GIT (B/16)~\citep{Wang2022GIT}, and Qwen2-VL-7B-Instruct~\citep{wang2024qwen2vlenhancingvisionlanguagemodels}. We prompt the models by asking: \textit{``What color is the word in this image?''} This phrasing is intentionally underspecified: it can be read as asking for the ink color or for the color named by the written word, mirroring the ambiguity that humans experience in the classical Stroop task.

Because the models generate free-form text, their response formats vary across models. Some models name the ink color, some repeat the written word, some explicitly mention both cues (e.g., ``The word 
\textsc{blue} is written in red''), and some drift off-target. We map each generated response into one of four mutually exclusive categories: \emph{Ink-only} when the model reports only the physical 
ink color, \emph{Word-only} when it reports only the written word's color identity, \emph{Both} when the response explicitly names both the written word's semantic color and the physical ink color as distinct attributes, and \emph{Neither} otherwise. Before 
categorization, color synonyms are normalized to the ten canonical labels (e.g., \emph{scarlet}\,$\rightarrow$\,\emph{red}); a response that maps to a single label after normalization is never assigned to 
\emph{Both}. All models are evaluated on the same 100-image Stroop dataset (10 congruent, 90 incongruent). Since the instruction explicitly asks for the ink color, we treat \emph{Ink only} as correct and \emph{Both} as ink-containing, as both include the 
task-relevant attribute. Exact label mappings are in 
Appendix~\ref{app:label_mapping}.

Similar to contrastive VLM, generative VLMs answered correctly on congruent images, where the ink color and written word matched (Figure~\ref{fig:generative_vlm_summary}C; the exceptions---GIT (80\%) and LLaVA (90\%)---reflect off-target outputs unrelated to the Stroop conflict, as confirmed by their \emph{Neither} counts. On incongruent images, however, performance 
dropped markedly for most models. Although the prompt asked \textit{``What color is the word in this image?''}, models often reported the semantic color of the written word rather than the physical ink color. Thus, conflicting textual information impairs the models' ability to report the task-relevant visual attribute. 

The degree of interference differed across models. Kosmos-2 exhibited the strongest word bias, 
consistently describing the written word and never reporting the ink color.  InstructBLIP occasionally reported the ink color (28.9\%), while BLIP-2 frequently produced responses mentioning both cues (31.1\% of incongruent trials; e.g., \textit{``The word \textsc{blue} is written in red''}), yet still showed a clear tendency to include the written word. GIT appeared strongly disrupted by the conflict: its responses were split between ink-color and word-based answers, with a sizable fraction classified as \emph{Neither} (18.9\%). LLaVA was the only model that reported the ink color more often than the word, but even so, ink-aligned responses occurred in only 54.4\% of incongruent trials.

Overall, generative VLMs are strongly influenced by conflicting textual cues: when word meaning and ink color disagree, and when the model are asked about the color of the word, most models fail to report the color and respond to the written word instead.

\textbf{Scale and Instruction Tuning Mitigates Cue Conflict}
Among the six models, \textbf{Qwen2-VL-7B-Instruct}~\citep{wang2024qwen2vlenhancingvisionlanguagemodels} stands out as the largest and most heavily instruction-tuned, with multimodal pretraining and alignment. To test whether scale and instruction tuning mitigate Stroop-like interference, we evaluated \textbf{Qwen2-VL-7B-Instruct} under two prompt formulations. The model 
answered the ink-color question perfectly on congruent images. On incongruent images, however, performance was strongly prompt-dependent. With the longer instruction: \emph{``You will see a single English word rendered in a colored ink. Ignore the written word and answer ONLY the 
ink color as one lowercase color name,''}, the model achieved 31.1\% ink-color accuracy. With the shorter instruction: \emph{``What is the ink color of the text in the image? Answer with one lowercase color word only,''} accuracy increased to 60.0\%. Counterintuitively, explicitly instructing the model to ignore the written word yielded \emph{lower} accuracy than the shorter prompt that does not mention 
the word at all---possibly because naming the word in the instruction primes the model to attend to it, amplifying rather than suppressing the interference. These results suggests that scale and instruction 
tuning reduce but do not eliminate feature interference, and the generated response are sensitive to prompt formulation. To directly address whether changing the prompt framing alters the results, we tested nine additional prompt variants in Appendix.
Overall, we observe that the response accuracy to ink color ranges from 24\% to 97\%, confirming that Stroop interference persists across prompt formulations, but its magnitude is sensitive to prompt specificity.

\begin{figure*}[t]
    \centering
    \includegraphics[width=0.75\textwidth]{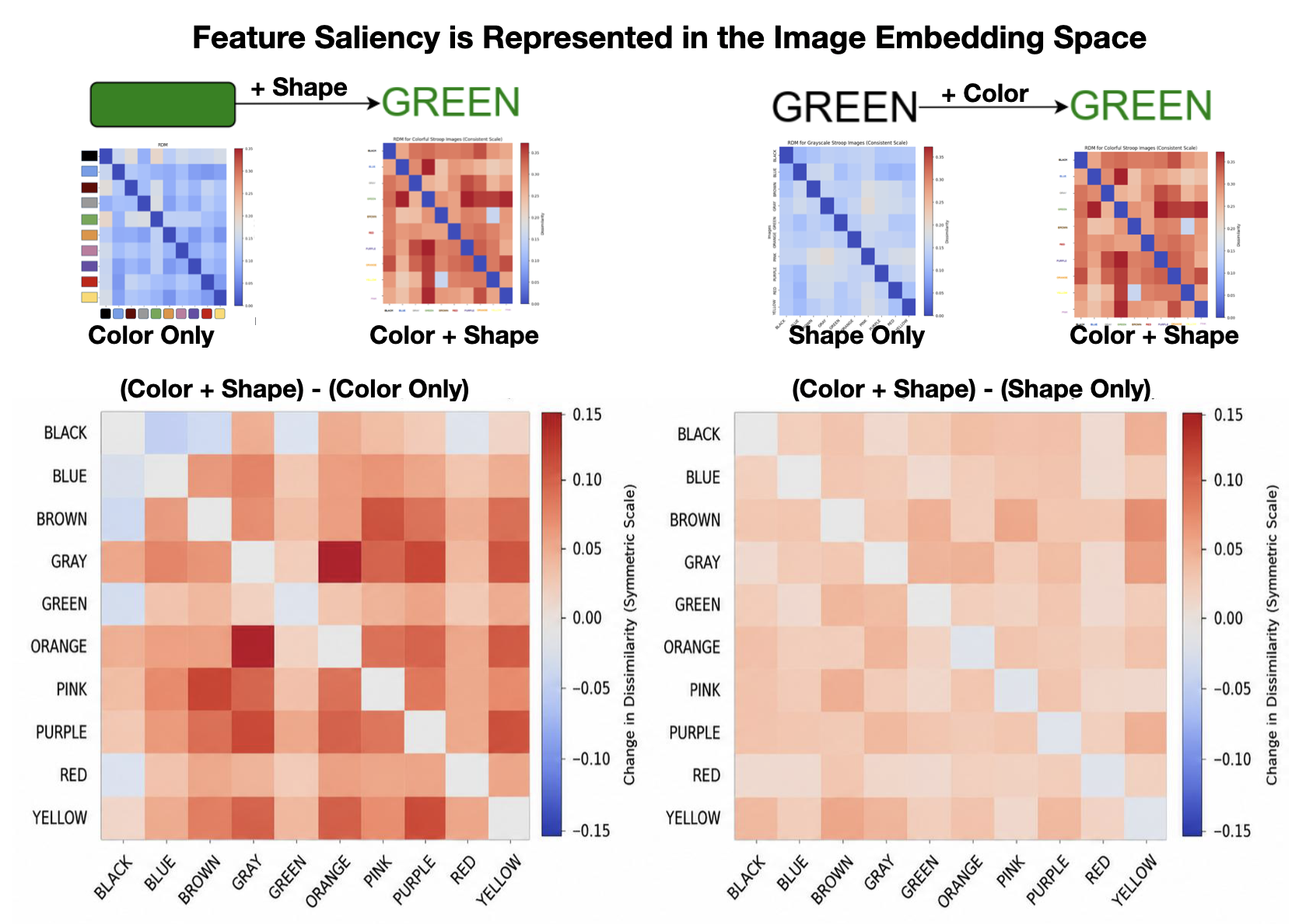}
    \caption{\textbf{Feature Saliency is Represented in the Image Embedding Space}
     Top: We calculate representation dissimilarity matrix among image embeddings of images that only contain colors \emph{ink-only}; images that only contain words (\emph{word-only}) images that contained colored words (\emph{word+ink}).
Bottom: resulting delta RDMs showing that the feature of word shape on top of colored feature makes image embeddings more dissimilar in than adding ink color feature on top of word shape. }
    \label{fig:featuresaliency}
\end{figure*}

\section{The Representation of Word Shape and Ink Color in the Image Embedding Space}
\label{sec:latent_intervention}

We next ask whether the VLM’s bias to word shape than to ink color is represented in the image embedding space. We collected CLIP's image embedding vectors for three groups of images: 
\emph{ink-only} (images of solid color without text), 
\emph{word-only} (grayscale words without color), 
and \emph{word+ink} (Stroop images containing colored words). 

We then analyzed the image embedding vectors using Representational Similarity Analysis (RSA), an approach from 
neuroscience ~\citep{Kriegeskorte2008} that compares similarities among high dimensional vectors of neural activities in response to groups of stimuli. RSA computes representational Dissimilarity Matrix (RDM), which are the dissimilarities between pairs of embeddings. Given image embeddings pairs $e_i$ and $e_j$ in response to an image pair, each entry of the RDM is defined as their cosine dissimilarity: 
$\text{RDM}(i,j) = 1 - \frac{e_i \cdot e_j}{\|e_i\|\|e_j\|}.$ 
Lower values indicate that the model represents the two stimuli more similarly in its embedding. 

We compute three RDMs separately for CLIP's embedding in response to the three groups of images. For each condition, we extract CLIP image embeddings for all 10 stimuli 
and compute a $10\times10$ pairwise cosine dissimilarity matrix.

To isolate the contribution of each feature (word shape or text color), we subtract RDMs:
\[
\Delta\text{Word} = \text{RDM}_{\text{Word+Ink}} - \text{RDM}_{\text{Ink-only}},
\]
\[
\Delta\text{Ink}  = \text{RDM}_{\text{Word+Ink}} - \text{RDM}_{\text{Word-only}}.
\]
$\Delta\mathrm{Word}$ measures how much the dissimilarity between two images changes when written words are added to images containing only ink colors. $\Delta\mathrm{Ink}$ measures the corresponding change when ink colors are added to images containing only written words. A large positive value means that adding the feature makes that pair of images more dissimilar in the model’s embedding space.

As shown in Figure~\ref{fig:featuresaliency}, $\Delta\mathrm{Word}$ contains larger, more concentrated positive regions: adding word feature makes image pairs much more dissimilar than they are when only ink color is present.  In contrast, $\Delta\mathrm{Ink}$ shows smaller, more diffuse changes: adding ink color to words shifts the embedding less. The RDM analyses suggest that the word feature is more salient than color 
More illustrations and discussions about the RDM analyses is included in Appendix.

\section{Steering the Concept Representations}
\begin{figure*}[htbp]
    \centering
    \includegraphics[width=0.8\textwidth]{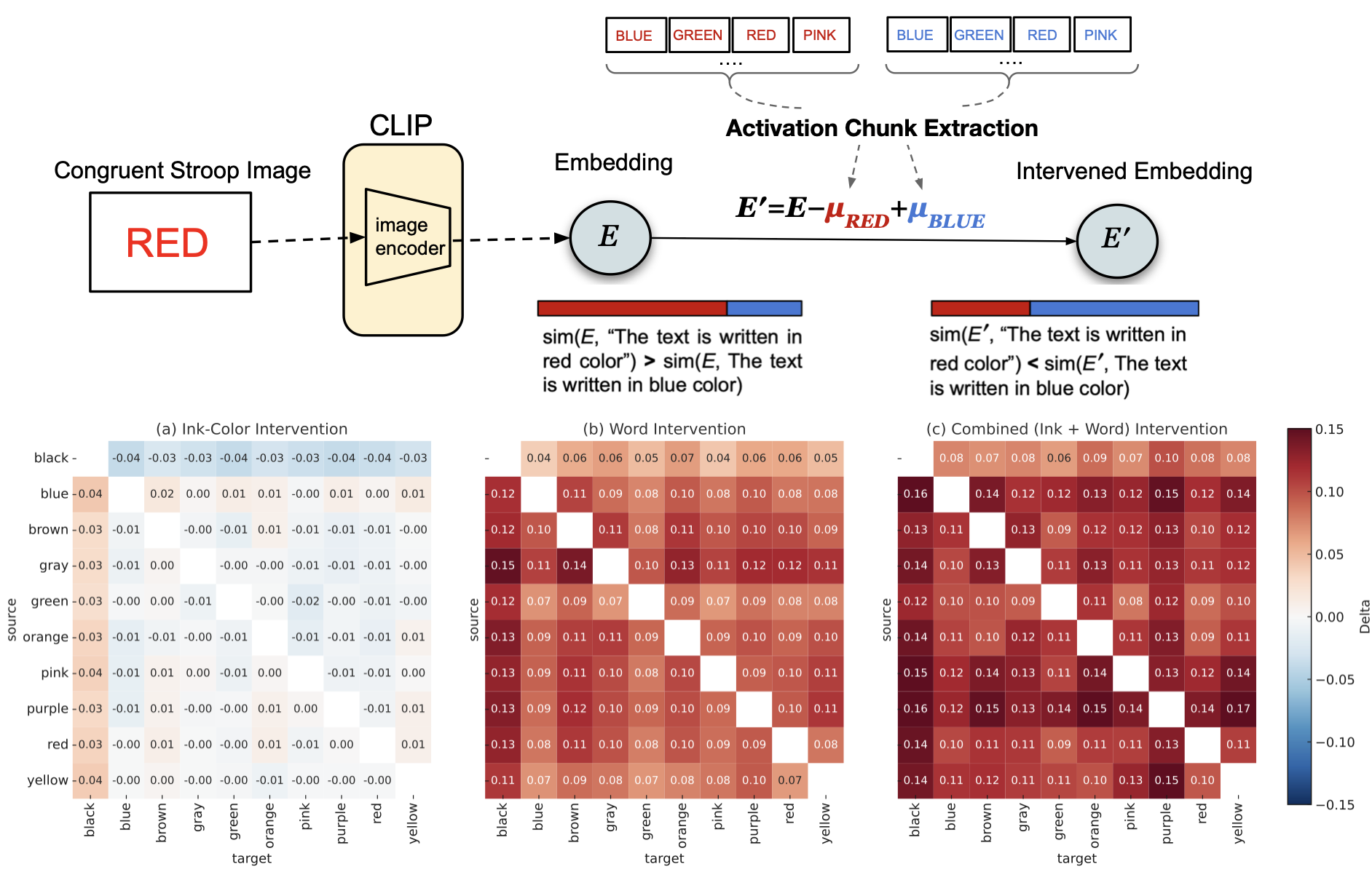}
    \caption{\textbf{Model Steering Pipeline} 
    A Stroop stimulus (e.g., the word ``\texttt{RED}'' in red ink) is encoded by CLIP into an embedding $E$. 
    We extract source ($\mu_{\text{src}}$) and target ($\mu_{\text{tgt}}$) encoding vectors using the chunk-based interpretability method by by aggregating embeddings from all images containing the concept (for colors, all items written in that ink color; for words, all items containing that word). 
     We then apply the intervention on the embedding vector $E$ as $E' = E - \mu_{\text{src}} + \mu_{\text{tgt}}$, and then re-evaluated the cosine similarity with source and target prompts.}
    \label{fig:intervention_pipeline}
\end{figure*}

We proceed to study whether manipulating image embeddings can change VLM's response tendency. To identify the dimensions that encode each concept, we use the chunk-based interpretability method of \citet{wu2025discoveringchunksneuralembeddings} to find the concept-encoding activations vectors. To do this, we collect embeddings from all images containing the concept. E.g. using all images written in red, regardless of the written word \emph{red ink} to extract representations of red concept. We then identify a subset of embedding dimensions $\Omega$ that respond consistently to that concept and compute its activation magnitude within this subspace (Figure \ref{fig:intervention_pipeline}) using the population averaging method \citet{wu2025discoveringchunksneuralembeddings}.

Denote $\mu_{\Omega(\mathrm{src})}$ and $\mu_{\Omega(\mathrm{tgt})}$ as the feature-encoding chunks of a source and target concept. Given the embedding $E$ of a image, we steer the model by subtracting the the source concept and add the target concept:
\[E'_{\Omega}=
E_{\Omega}
-
\mu_{\Omega(\mathrm{src})}
+
\mu_{\Omega(\mathrm{tgt})}.
\]

We consider three interventions: \emph{ink-color steering} (e.g., red $\rightarrow$ blue), \emph{word steering} (e.g., \texttt{RED} $\rightarrow$ \texttt{BLUE}), and \emph{combined steering}, in which both shifts are applied. 

We compare the steered image embedding $E'_{\Omega}$ with text embedding the targeted and source concepts for each intervention:
\[
\Delta
=
\operatorname{sim}(E',p_{\mathrm{tgt}})
-
\operatorname{sim}(E',p_{\mathrm{src}}).
\]
An intervention is successful when $\Delta>0$: the edited image embedding is more similar to the targeted text embedding than to the text embedding of the source prompt. 



We observed that the representation for word features are highly steerable: word interventions succeed in \textbf{100\%} of source--target pairs, with a mean similarity shift of $+0.0934 \pm 0.0214$. In contrast, steering fetures that encode ink-color interventions succeeded in only \textbf{36.67\%} of cases and produce essentially no average shift ($-0.0017 \pm 0.0166$) across the colors. Combined word-and-color interventions also achieve \textbf{100\%} success, with an even larger mean shift of $+0.1172 \pm 0.0215$. 

This asymmetry is consistent with the underlying geometry of the concept-encoding vectors. Word vectors have substantially larger $\ell_2$ norms (mean $6.36$) and are more distinct from one another, whereas ink-color vectors are smaller (mean $2.99$) and strongly collinear (mean pairwise cosine similarity $\approx 0.79$). Word feature is represented by stronger and more separable directions in CLIP embedding space than ink color.

Figure~\ref{fig:intervention_pipeline} shows $\Delta$ for each source--target pair under Ink, Word, and Combined interventions. Positive values indicate successful steering toward the target concept. Word and Combined interventions produce consistently positive shifts across source--target pairs, whereas Ink-Color interventions remain weak and diffuse. 


\section{Beyond Text and Color}
\label{sec:flux_realistic}

To explore cue conflicts in a more general and realistic setting, we generated a set of realistic multimodal conflict images using the FLUX.1 model, including contradictory cases such as a green ``STOP'' sign or a red Wi-Fi icon labeled ``CONNECTED''. For each image, we paired a \emph{word prompt} 
(describing the written content) with a \emph{color prompt} (describing the dominant visual cue), and 
measured CLIP’s similarity to determine whether it favored the textual or visual signal. 

Across these stimuli, CLIP alternated between word-based and color-based decisions depending on cue 
saliency: text-heavy signage induced word dominance, whereas icon-like or strongly chromatic regions 
elicited color dominance. Two examples are shown in Table~\ref{tab:flux_examples}; the full set with 
similarity scores and interpretations is provided in Appendix~\ref{app:flux_examples}.

\begin{table}[h!]
\centering
\footnotesize
\caption{\textbf{Illustrative FLUX-generated conflict cases.} 
CLIP’s decision reflects either the textual or visual cue, depending on saliency.}
\label{tab:flux_examples}
\setlength{\tabcolsep}{3pt}
\begin{tabular}{c p{2.8cm} c c c}
\toprule
\textbf{Image} & \textbf{Visual Description} & \textbf{word\_sim} & \textbf{color\_sim} & \textbf{Decision} \\
\midrule
\raisebox{-0.5\height}{\includegraphics[width=0.12\columnwidth]{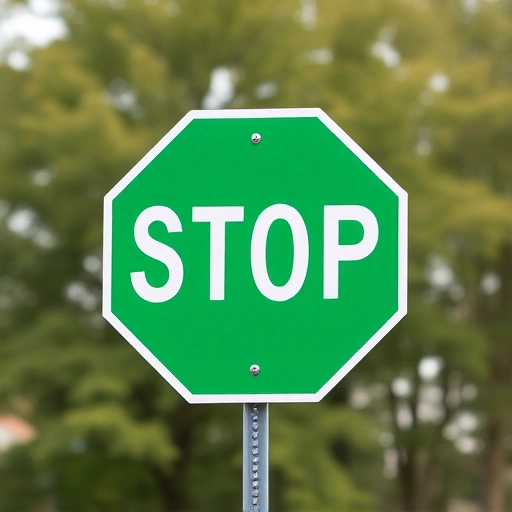}} 
& ``STOP'' text in green sign 
& 0.17 & 0.83 & COLOR \\
\midrule
\raisebox{-0.5\height}{\includegraphics[width=0.12\columnwidth]{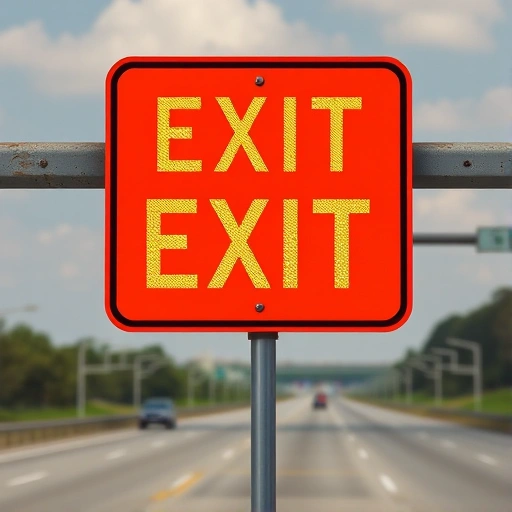}} 
& ``EXIT'' (red sign) 
& 0.99 &0.001 & WORD \\
\bottomrule
\end{tabular}
\end{table}

\section{Conclusion}
\label{sec:conclusion}
In this work, we adapt the Stroop paradigm and systematically test Vision--Language Models' behavior in the presence of conflicting cues between the \emph{word} (the printed string) and the \emph{ink color}. We observed a consistent behavioral bias across contrastive encoders (CLIP, SigLIP-2) and generative VLMs (BLIP-2, InstructBLIP, Kosmos-2, LLaVA, GIT, Qwen2-VL): when word and ink color disagree, models overwhelmingly align their preferences with the word, similar to humans. This tendency persists despite varying font size, font weight, and contrast. Going beyond behavioral observations, we analyzed CLIP’s embedding space and found that neural subpopulation vectors encoding words are long and modular, whereas those encoding ink-color are short and collinear. This representation saliency is reflected in steerability: steering embedding space towards word directions consistently flipped the network’s preference, while steering towards ink-color direction is less effective. This suggests that word over color dominance is represented in the embedding space. Together, by adapting a classic psychological paradigm to VLMs, our study shows that models prioritize word over color cues when they conflict, a bias rooted in unequal embedding representations. 
Our systematic study on multi-cue conflict in the Stroop test provides a foundation for future understanding of cue representation in the most general setting.

\section{Discussion and Limitations}
\label{sec:discussion_limitations}
Our study has limitations. Due to limited computational resources, embedding-level interventions were conducted only in CLIP. Moreover, our categorization of generative VLM responses 
(\emph{Word-only}, \emph{Ink-only}, \emph{Both}, \emph{Neither}) may miss edge cases, for instance, when models produce descriptive sentences mixing both attributes. Additionally,  chunk extraction for ink color likely underestimates steerability if color is more distributed than our centroids capture, and SigLIP-2’s rigidity under legibility manipulations suggests model-specific sensitivities that we did not probe exhaustively. Future work may extend latent interventions beyond CLIP to recent VLMs (e.g., LLaVA-1.6, InternVL-2.5, Qwen-2.5-VL, SigLIP-2), as well as to multi-image and video inputs. Another direction is to move from synthetic stimuli to more naturalistic conflicts involving textures, materials, and layered graphics. Finally, future work could develop stronger ink/color representations in the embedding space, for example by improving chunking methods, using training data that emphasizes color independently of semantics, or exploring architectural approaches that explicitly decouple reading from seeing. 
Prior work has documented broader forms of textual dominance 
\citep{menon2022taskbiasvisionlanguagemodels,pezeshkpour2025mixedsignalsdecodingvlms,deng2025wordsvisionvisionlanguagemodels} 
and difficulty integrating conflicting visual–textual cues 
\citep{jia2025benchmarkingmultimodalknowledgeconflict}, 
as well as color-processing weaknesses in contrastive encoders 
\citep{Arias_2024}. 
However, these analyses do not isolate the specific, tightly-controlled word–vs–color conflict we study, nor do they connect behavioral asymmetry to representational geometry or steerability. Our Stroop-style paradigm addresses this gap by providing a mechanistic explanation of why word cues dominate.

Furthermore, future work may examine scaling effects within the same model family(e.g., Qwen2-VL 7B → 13B) to test whether the word–color asymmetry weakens with capacity. A complementary direction is to analyze token-probability dynamics, which may reveal Stroop-like slowdowns in model confidence that are out of scope for this study.

\bibliography{aaai2027}

\appendix

\section{Stroop Dataset and Prompt Design}
\label{appendix:dataset}
To investigate how Vision–Language Models (VLMs) handle multimodal conflicts, we constructed a synthetic Stroop-style dataset.

\subsection{Dataset Overview}
The dataset consists of \textbf{100 Stroop stimuli} generated programmatically. Each image contains a single uppercase color word rendered in a specific ink color.
\begin{itemize}
    \item \textbf{Color vocabulary (10 classes):} \textit{red, blue, green, yellow, orange, purple, brown, pink, gray, black}.
    \item \textbf{Stimulus types:} \textbf{10} congruent (word==ink) and \textbf{90} incongruent (word$\neq$ink).
    \item \textbf{Image design:} white background, centralized positioning, uniform font (90\,pt).
    \item \textbf{Resolution \& format:} 256$\times$256 px PNG (sRGB, embedded ICC profile).
\end{itemize}
Each of the ten words appears exactly once in each of the ten ink colors, yielding a set of \textbf{100} stimuli in total (\textbf{10 congruent}, \textbf{90 incongruent}). Every color word is rendered once in its matching ink and nine times in non-matching inks, and the same symmetry holds when counted by ink color.

\subsection{Color Specification}
We fix sRGB hex values to avoid palette drift and to ease reproduction.

\begin{table*}[htbp]
\centering
\resizebox{\textwidth}{!}{%
\begin{tabular}{lcccccccccc}
\toprule
Color  & red & blue & green & yellow & orange & purple & brown & pink & gray & black \\
\midrule
Hex    & \texttt{fc0404} & \texttt{0c0cfc} & \texttt{048404} & \texttt{fcfc17} & 
\texttt{fca404} & \texttt{8c178c} & \texttt{8c4414} & \texttt{fcc4cc} & 
\texttt{848484} & \texttt{000000} \\
\bottomrule
\end{tabular}}
\caption{Fixed sRGB hex codes for ink colors.}
\label{tab:hex_codes}
\end{table*}

\newcommand{\swatch}[2]{%
  \begin{tabular}{@{}c@{}}
    \setlength{\fboxsep}{0pt}%
    \colorbox{#1}{\phantom{\rule{10mm}{7mm}}}\\[-0.6ex]
    \scriptsize #2
  \end{tabular}%
}

Each model was evaluated on all 100 Stroop images (10 congruent, 90 incongruent). For visual manipulation experiments (font size, weight, contrast, pseudowords), we generated an additional 400–500 samples per manipulation, resulting in a few thousand evaluation datapoints overall. 

\subsection{Evaluation Scale (Datapoints)}
\label{app:evaluation_scale}
\begin{table}[H]
\centering
\scriptsize
\setlength{\tabcolsep}{6pt}
\renewcommand{\arraystretch}{1.15}
\begin{tabular}{|l|c|c|c|}
\hline
\textbf{Experiment type} & \textbf{\#stimuli} & \textbf{\#models} & \textbf{Total evaluations} \\
\hline
Base Stroop (cong+incong) & 100  & 7               & 700   \\
Font size (4$\times$100)  & 400  & 2 (CLIP+SigLIP-2) & 800   \\
Font weight (4$\times$100)& 400  & 2               & 800   \\
Contrast (4$\times$100)   & 400  & 2               & 800   \\
Pseudowords               & 500  & 2               & 1000  \\
\rowcolor{gray!10}
Qwen2-VL (2 prompts)      & 100  & 1               & 200   \\
\hline
\textbf{Total}            & --   & --              & $\sim$4{,}300 \\
\hline
\end{tabular}
\caption{Scale of evaluation datapoints across experiments (updated with Qwen2-VL).}
\label{tab:evaluation_scale}
\end{table}

\subsection{Prompt Design}
We use two prompt families for CLIP and a single standardized question for generative VLMs.

\begin{itemize}
    \item \textbf{Ink-oriented (visual):} ``The text is written in red color.''
    \item \textbf{Word-oriented (semantic):} ``The text says \textsc{RED}.''
\end{itemize}

\noindent\textbf{Generative VLM query:}
\begin{quote}
\textit{``What color is the word in this image?''}
\end{quote}

\noindent For Qwen2-VL-7B-Instruct we used two prompt variants; see App.~\ref{app:qwen}.

\subsection{Visual Manipulations}
We additionally probe legibility and perceptual robustness via controlled perturbations.

\begin{table*}[H]
\centering
\begin{tabular}{|l|p{7cm}|l|}
\hline
\textbf{Manipulation} & \textbf{Description} & \textbf{Purpose} \\
\hline
Font size & 48, 72, 90, 108\,pt & Visual salience / overflow check \\
Font weight & Bold, Normal, Light, Narrow & Readability vs.\ ink reliance \\
Tone variation & 10 brightness/saturation levels  & Perceptual robustness \\
Contrast & High / Medium / Low / Same (text color equals background) & Visibility thresholding \\
Pseudowords & Word replaced by nonwords (e.g., \texttt{GNEK}, \texttt{GRIV}, \texttt{GRUP}, ...) & Remove lexical semantics \\
\hline
\end{tabular}
\caption{Visual manipulations applied to Stroop stimuli.}
\label{tab:visual_manipulations}
\end{table*}

\subsection{Example Stimuli}
\begin{figure}[H]
\centering
\includegraphics[width=\columnwidth]{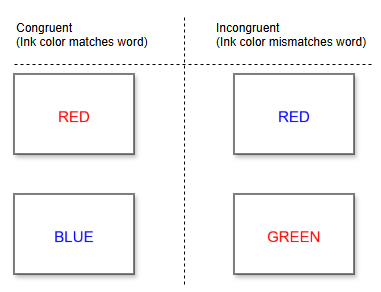}
\caption{Examples from the dataset showing congruent and incongruent stimuli.}
\label{fig:appendix_stroop_examples}
\end{figure}

\section{Full Behavioral Results}
\label{app:behavioral}

\subsection*{ CLIP Behavioral Results}



\paragraph{Family-specific accuracies (for comparability with the main text)}
For completeness, we also report accuracies \emph{within} each prompt family, i.e., scoring the image against the color prompts only, or against the text prompts only. These are the numbers cited in the main paper.

\begin{table*}[!htbp]
\centering
\begin{tabular}{|l|c|c|}
\hline
\textbf{Prompt family} & \textbf{Congruent (n=10)} & \textbf{Incongruent (n=90)} \\
\hline
Word-oriented (\textit{``The text says X''})      & 10/10 (100\%) & 88/90 (\textbf{97.8\%}) \\
Ink-oriented (\textit{``Written in X color''})  & 10/10 (100\%) & 18/90 (\textbf{20.0\%}) \\
\hline
\end{tabular}
\caption{Family-specific accuracies. Reporting both views avoids conflating the \emph{decision rule} (winner-takes-all) with \emph{family-specific} correctness.}
\label{tab:clip_family_specific}
\end{table*}
\paragraph*{Visual Manipulation Effects}

\begin{figure}[h!]
  \centering
  \includegraphics[width=\columnwidth]{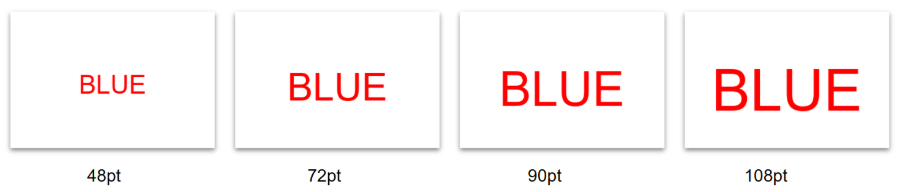}
  \caption{Font size exemplars and aggregate trend (Table~\ref{tab:manip_summary}).}
  \label{fig:clip_visual_fontsize}
\end{figure}

\begin{figure}[h!]
  \centering
  \includegraphics[width=\columnwidth]{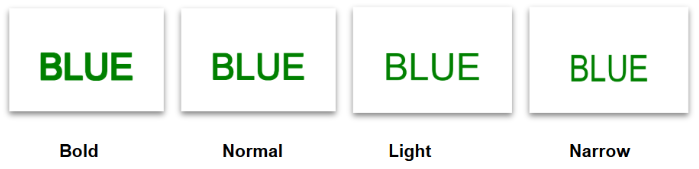}
  \caption{Font weight exemplars and aggregate trend (Table~\ref{tab:manip_summary}).}
  \label{fig:clip_visual_boldness}
\end{figure}

\begin{figure}[h!]
  \centering
  \includegraphics[width=\columnwidth]{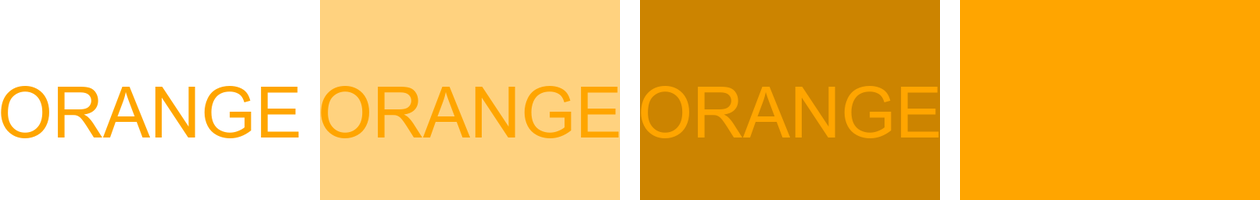}
  \caption{Contrast manipulation exemplars; see Table~\ref{tab:manip_summary}.}
  \label{fig:clip_visual_contrast}
\end{figure}

\begin{figure}[h!]
  \centering
  \includegraphics[width=\columnwidth]{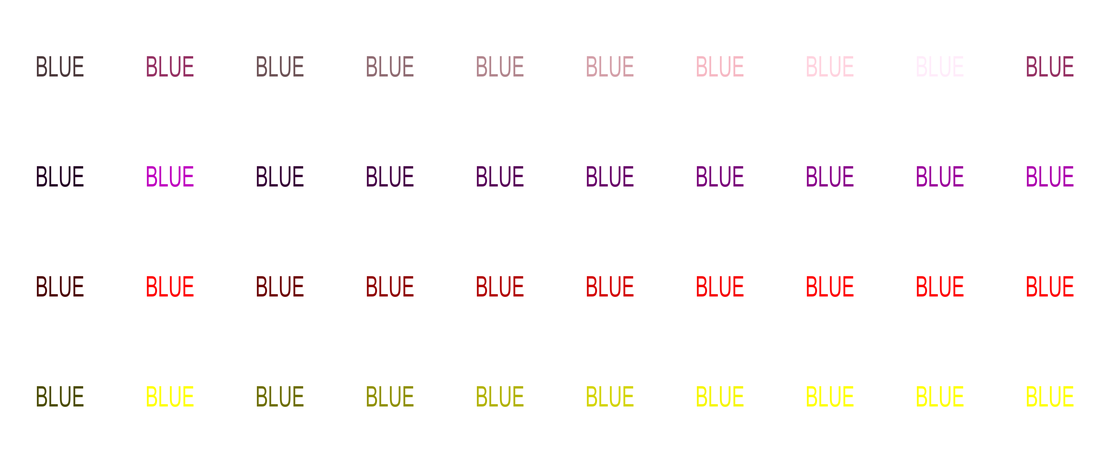}
  \caption{Tone variations (examples).}
  \label{fig:clip_visual_background}
\end{figure}

\begin{figure}[h!]
  \centering
  \includegraphics[width=\columnwidth]{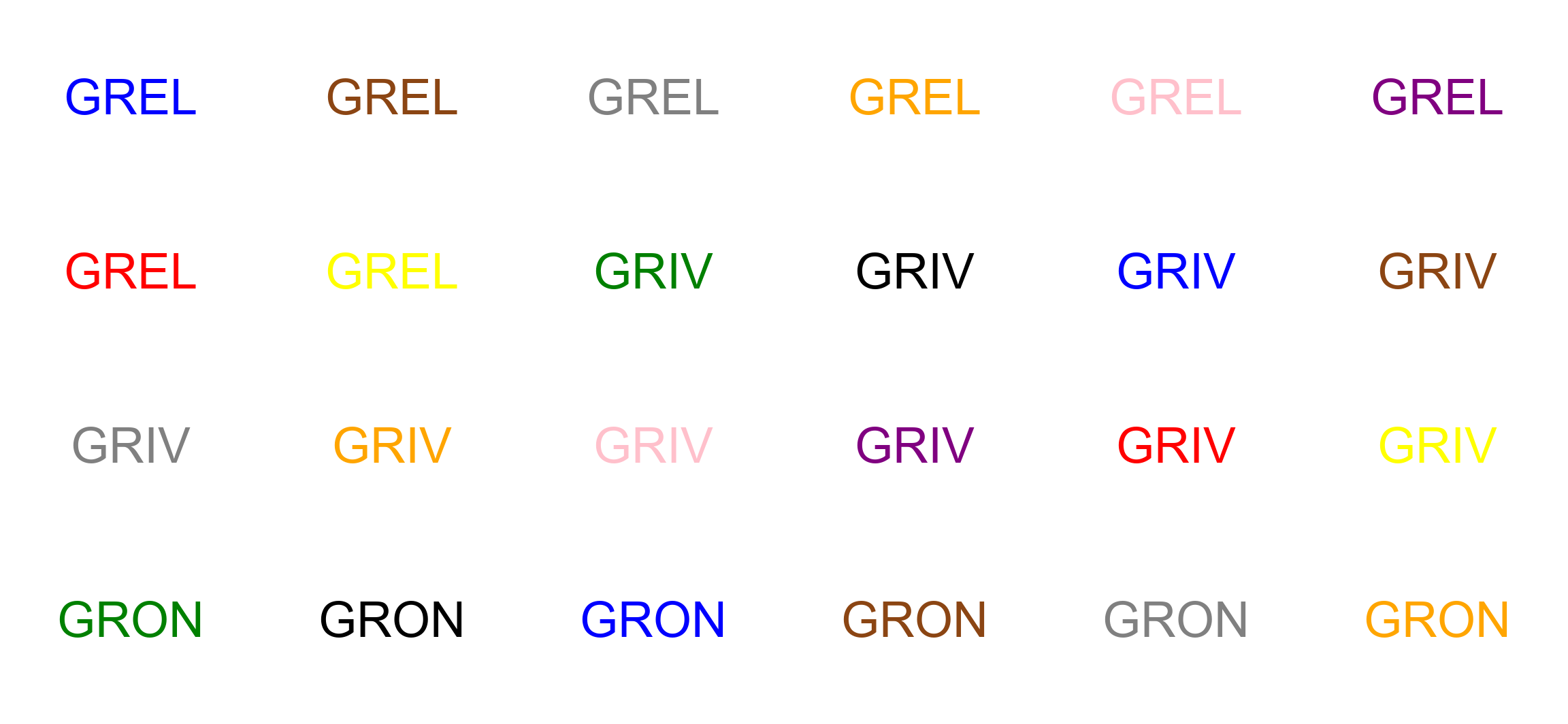}
  \caption{Pseudoword exemplars; no lexical cue leads to color-based decisions.}
  \label{fig:clip_visual_pseudo}
\end{figure}

We probe whether changing legibility or perceptual salience shifts CLIP’s preference. Four manipulations are considered: font size, font weight, contrast, and pseudowords. Figures~\ref{fig:clip_visual_fontsize}--\ref{fig:clip_visual_pseudo} show exemplars; aggregated numbers are summarized below.

\begin{table*}[!htbp]
\centering
\scriptsize 
\setlength{\tabcolsep}{4pt} 
\renewcommand{\arraystretch}{1.1} 
\begin{tabular}{|l|c|c|c|c|}
\hline
\textbf{Manipulation} & \textbf{Setting} & \textbf{Text match} & \textbf{Color match} & \textbf{Notes} \\
\hline
Font size & 48\,pt  & 97\% & 13\% & Readable; text dominates \\
          & 72\,pt  & 95\% & 15\% & \\
          & 90\,pt  & 89\% & 21\% & \\
          & 108\,pt & 77\% & 33\% & Overflow reduces readability \\
\hline
Font weight & Bold   & 89\% & 21\% & Minor effect on color \\
            & Normal & 88\% & 22\% & \\
            & Light  & 89\% & 21\% & \\
            & Narrow & 97\% & 13\% & Highest text bias \\
\hline
Contrast (incongruent) & High   & 87.78\% & 12.22\% & \multirow{4}{*}{Shift from text$\rightarrow$color as contrast drops} \\
                       & Medium & 83.33\% & 16.67\% & \\
                       & Low    & 67.78\% & 32.22\% & \\
                       & Same   & 0.00\%  & 100.00\% & Text unreadable \\
\hline
Tone variation (incongruent) & All tones & 86.40\% & 3.60\% & Lower chroma reduces color salience; text dominates \\
\hline
Pseudowords (incongruent) & — & \multicolumn{2}{c|}{Color match $\approx$ 78\%} & No lexical cue; color used \\
\hline
\end{tabular}
\caption{Effect of visual manipulations (aggregated across colors). Values mirror the panels in Figs.~\ref{fig:clip_visual_fontsize}--\ref{fig:clip_visual_pseudo}; tone-variation exemplars are in Fig.~\ref{fig:clip_visual_background}.}
\label{tab:manip_summary}
\end{table*}

\noindent
 CLIP “reads when it can”: high legibility (smaller size, heavier weight, strong contrast) reliably yields text-aligned decisions; suppressing legibility (overflow, low/same contrast, pseudowords) flips decisions toward ink color. The trend is monotonic under contrast (High$\to$Same) and only partially attenuated by making text visually larger/heavier.


\paragraph*{Pseudoword Response Categories}

Since pseudowords carry no semantic meaning, strictly speaking, there are no true ``congruent'' cases. 
We therefore break down responses into four categories:

\begin{itemize}
    \item \textbf{Color Match (ink-based):} The model outputs the correct ink color, independent of the pseudoword form. 
    Example: \textit{GRIV} printed in green ink, model answers ``green.''

    \item \textbf{Form–Color Aligned:} In some cases, a pseudoword resembles a real color word (e.g., \textit{BLIR} $\sim$ ``blue''). 
    When printed in that same color, both cues align. Earlier drafts labeled this as ``congruent,'' 
    but here we adopt the more precise term \emph{Form–Color Aligned}.

    \item \textbf{Text-Biased (form-based):} The model treats the pseudoword as if it were a real color word, even when the ink color differs. 
    Example: \textit{GRIV} printed in red ink, model answers ``green.''

    \item \textbf{Other Errors:} Predictions that match neither ink color nor pseudoword form. 
    Example: \textit{GRIV} printed in red ink, model answers ``purple.''
\end{itemize}

\begin{table}[!htbp]
\centering
\scriptsize
\begin{tabular}{lcc}
\toprule
\textbf{Category} & \textbf{Count} & \textbf{Accuracy (\%)} \\
\midrule
Color Match (ink-based) & 389 & 77.8 \\
Form–Color Aligned & 49 & 9.8 \\
Text-Biased (form-based) & 28 & 5.6 \\
Other Errors & 34 & 6.8 \\
\midrule
\textbf{Total} & 500 & 100 \\
\bottomrule
\end{tabular}
\caption{Aggregated pseudoword results across all colors ($N=500$). 
No true ``congruent'' cases exist; \textbf{Form–Color Aligned} refers to coincidental alignment of pseudoword form and ink color.}
\label{tab:pseudoword_agg}
\end{table}

\paragraph*{Nonsense-Word Control Experiment (“zarp”)}

To clarify whether CLIP's behavior in the Stroop setting reflects 
genuine color understanding or simply an OCR-driven preference for 
whatever text appears in the image, we conducted a nonsense-word 
control experiment using the pseudoword \textit{zarp}. We 
regenerated the entire Stroop set with the word \textit{zarp} 
rendered in each of the ten ink colors, and—critically—added 
\textit{zarp} to the list of candidate color labels so that the 
model could select it as a legitimate answer.

The outcome was unambiguous: across all ink colors, CLIP chose \textit{zarp} in 
\textbf{100\%} of trials, resulting in \textbf{0\%} ink accuracy. In other words, 
even when the printed token carries no meaning, CLIP treats it as a valid color 
category purely because it is visually present in the image. 

This result makes the underlying mechanism clear. When faced with an unfamiliar 
string, CLIP does not rely on color semantics at all; instead, it aligns the 
image embedding with the closest matching text embedding in a strongly 
OCR-like manner. 
The word-over-color effect does not reflect semantic reasoning about colors, 
but rather a structural bias in CLIP’s embedding geometry that privileges 
textual features over chromatic ones.

\subsection{SigLIP-2 Behavioral Results}
\label{app:siglip}

While SigLIP-2 introduces architectural improvements over CLIP, its behavior under multimodal conflict remains strongly word-dominant. Below we provide condition-wise breakdowns.

\paragraph{Congruent vs.\ Incongruent Cases}
\begin{table}[H]
\centering
\begin{tabular}{|l|c|c|}
\hline
\textbf{Condition} & \textbf{Text Accuracy (\%)} & \textbf{Color Accuracy (\%)} \\
\hline
Congruent & 100.0 & 100.0 \\
Incongruent & 100.0 & 5.6 \\
\hline
\end{tabular}
\caption{SigLIP-2 behavioral accuracy on congruent and incongruent Stroop stimuli.}
\label{tab:siglip_basic}
\end{table}

\paragraph{Font Size Variation (48–108 pt)}
\begin{table}[!htbp]
\centering
\begin{tabular}{|l|c|c|}
\hline
\textbf{Font Size} & \textbf{Text Accuracy (\%)} & \textbf{Color Accuracy (\%)} \\
\hline
48\,pt  & 100.0 & 15.0 \\
72\,pt  & 100.0 & 16.0 \\
90\,pt  & 100.0 & 16.0 \\
108\,pt & 100.0 & 16.0 \\
\hline
\end{tabular}
\caption{Accuracy by font size for SigLIP-2. Unlike CLIP, larger sizes did not weaken text dominance.}
\label{tab:siglip_fontsize}
\end{table}

\paragraph{Font Weight Variation}
\begin{table}[!htbp]
\centering
\begin{tabular}{|l|c|c|}
\hline
\textbf{Style} & \textbf{Text Accuracy (\%)} & \textbf{Color Accuracy (\%)} \\
\hline
Light  & 100.0 & 6.7 \\
Normal & 100.0 & 6.7 \\
Bold   & 100.0 & 7.8 \\
Narrow & 100.0 & 6.7 \\
\hline
\end{tabular}
\caption{Accuracy by font weight. SigLIP-2 remained insensitive to style; text was always dominant.}
\label{tab:siglip_fontweight}
\end{table}

\paragraph{Contrast Manipulation}
\begin{table}[H]
\centering
\begin{tabular}{|l|c|c|c|}
\hline
\textbf{Contrast} & \textbf{Text Win (\%)} & \textbf{Color Win (\%)} & \textbf{Tie (\%)} \\
\hline
High   & 55.6 & 11.1 & 33.3 \\
Medium & 44.4 & 11.1 & 44.4 \\
Low    & 22.2 & 66.7 & 11.1 \\
Same   & 33.3 & 11.1 & 55.6 \\
\hline
\end{tabular}
\caption{Preference breakdown for contrast manipulations. SigLIP-2 showed indecision (ties) at medium/same levels, unlike CLIP which switched fully to color when text was unreadable.}
\label{tab:siglip_contrast}
\end{table}

\paragraph{Pseudowords}
\begin{table}[!htbp]
\centering
\begin{tabular}{|l|c|}
\hline
\textbf{Outcome} & \textbf{Count (\%)} \\
\hline
Ink-oriented prompt wins & 335 (67\%) \\
Word-oriented prompt wins  & 139 (28\%) \\
Tie                        & 26  (5\%) \\
\hline
\end{tabular}
\caption{SigLIP-2 on pseudoword stimuli. Unlike CLIP, which often clung to text shapes, SigLIP-2 switched reliably to color.}
\label{tab:siglip_pseudoword}
\end{table}

Across all manipulations, SigLIP-2 exhibited a rigid Stroop-like bias toward text, with 100\% text accuracy under incongruence. Visual manipulations (size, weight, contrast) did not weaken the bias, though pseudowords revealed a partial recovery of color accuracy. These results indicate that SigLIP-2, despite training improvements, inherits the same text-over-color preference as CLIP.

\subsection{Large-Scale Stroop Dataset (23k Images)}
\label{app:large_stroop}

To complement the controlled 100-image Stroop set, we generated a large-scale 
variant consisting of 23,338 images. While congruent cases were kept clean and 
minimal (10 total), the remaining 23,328 incongruent stimuli were produced using 
systematic variations in background texture, tone, saturation, brightness, and 
mild spatial perturbations. These manipulations ensure that models cannot rely on 
template memorization and instead must perform genuine color recognition under 
appearance diversity.

Each stimulus was rendered using the same 10 ink-color categories as the main 
dataset. For evaluation, CLIP was tested with the standard ink- and word-oriented 
prompt sets. The model maintained extremely high word accuracy (99.5\%) while ink 
accuracy remained low (15.9\%), reproducing the strong Stroop-style word bias 
observed in the main paper. Full sampling parameters, rendering functions, and the generation script are included in the project repository and will be released in a future public release.

\section{VLM Model Details}
\label{app:model_details}

Table~\ref{tab:model_details} summarizes the specific versions of all models used in our experiments, including their backbone, approximate parameter count, training objective, and release source. 
\begin{table*}[t]
\centering
\small
\setlength{\tabcolsep}{4pt}
\begin{tabular}{|p{2.4cm}|p{3.1cm}|p{1.3cm}|p{3.4cm}|p{2.4cm}|}
\hline
\textbf{Model} & \textbf{Backbone} & \textbf{Params} & \textbf{Training Objective} & \textbf{Source / Release} \\
\hline
CLIP & ViT-B/32 + Transformer text encoder & $\sim$150M & Contrastive (softmax) & \citet{Radford2021} \\
\hline
SigLIP-2 & ViT-B/16 (base) + text encoder & $\sim$150M & Contrastive (sigmoid) & Google, 2024 release \\
\hline
BLIP-2 & ViT-g + FlanT5-XL & $\sim$13B & Q-former + pretrain + instruction tuning & \citet{Li2022BLIP} \\
\hline
InstructBLIP & ViT-g + Vicuna-7B & $\sim$7B & Instruction-tuned generation & \citet{dai2023instructblip} \\
\hline
Kosmos-2 & Multi-stream transformer (1B) & $\sim$1B & Multimodal pretrain (text+image) & \citet{peng2024grounding} \\
\hline
LLaVA (v1.6) & CLIP ViT-L/14 + Vicuna-7B & $\sim$7B & Instruction-tuned generation & \citet{liu2023visualinstructiontuning} \\
\hline
GIT & ViT-B/16 encoder + GPT2-style decoder & $\sim$400M & Pretrain (captioning) & \citet{Wang2022GIT} \\
\hline
Qwen2-VL-7B-Instruct & ViT-based vision encoder + 7B LLM & $\sim$7B & Multimodal instruction tuning & \citet{wang2024qwen2vlenhancingvisionlanguagemodels} \\
\hline
\end{tabular}
\caption{Model configurations used in this study. Parameter counts are approximate; all models are publicly available.}
\label{tab:model_details}
\end{table*}

\section{Label Mapping and Post-processing Rules}
\label{app:label_mapping}

For evaluation, raw model outputs or similarity scores are mapped into four categories: 
\textbf{Word Match}, \textbf{Ink Match}, \textbf{Both}, and \textbf{Neither}. 
The mapping procedure ensures consistency across CLIP/SigLIP (similarity-based) and generative VLMs (free-form text outputs).

\begin{itemize}
    \item \textbf{Word-oriented prompts} (“The text says X”): If the top-1 prediction is X, we assign a \emph{Word Match}.
    \item \textbf{Ink-oriented prompts} (“The text is written in X color”): If the top-1 prediction is X, we assign an \emph{Ink Match}.
    \item \textbf{Tie-handling (similarity-based models)}: When cosine similarities are equal, ties are resolved by a fixed priority: Word Match $>$ Ink Match $>$ Neither. 
    \item \textbf{Generative VLM outputs}: Free-form text is normalized (lowercased, trimmed, and mapped to the set of 10 basic colors \{red, blue, green, yellow, orange, pink, purple, black, brown, gray\}). Non-matches are labeled as \emph{Neither}.
    \item \textbf{Qwen2-VL prompts:} responses are already constrained to a single lowercase color word; we still normalize to the 10-color lexicon and map out-of-vocabulary items to \emph{Neither}.

\end{itemize}

\subsection{Generative VLM Scoring Procedure (TSR)}
\label{app:tsr}

For completeness and to ensure comparability across all evaluated models, we 
report here the procedure used to score free-form outputs from generative 
VLMs. All responses are processed through a deterministic, rule-based 
normalization pipeline: outputs are lowercased, stripped of punctuation, 
and matched against a fixed lexicon of ten canonical color labels 
\{red, blue, green, yellow, orange, pink, purple, black, brown, gray\}. 
Responses that do not match any valid label are assigned the category 
\emph{Neither}. No LLM-as-a-Judge or heuristic prompt interpretation is used.

Because instruction-tuned generative models are expected to follow formatting 
constraints, we additionally compute a \textbf{Template Success Rate (TSR)}, 
defined as the fraction of outputs that adhere to the required format of 
a single lowercase color word. This allows us to separate genuine Stroop 
behavior from potential instruction-following failures.

\begin{itemize}
    \item \textbf{Qwen2-VL-7B-Instruct}: TSR = 100\%.  
    All responses were correctly formatted single-token color labels. 
    Nevertheless, decisions were strongly dominated by the written word 
    (97\% word-match vs.\ 13\% ink-match), indicating that the observed 
    Stroop pattern is not attributable to formatting or parsing issues.

    \item \textbf{LLaVA-1.5}: TSR = 92\%.  
    Among parseable outputs, the model exhibited a similar bias 
    (88\% word-match vs.\ 12\% ink-match), again showing that instruction 
    compliance does not eliminate word dominance.
\end{itemize}

For caption-driven or non-instruction-tuned models (e.g., BLIP-2, Kosmos-2), 
strict templating is not meaningful: these models are not designed to output 
single-token answers. Applying TSR would therefore conflate instruction 
capability with Stroop sensitivity. We instead evaluate such models using 
their unconstrained outputs passed through the same normalization and lexicon 
mapping pipeline described above.

Taken together, these results confirm that the Stroop-style word preference in 
generative VLMs persists even when instructions are fully obeyed and outputs 
are perfectly formatted. The bias therefore reflects a genuine property of 
their underlying semantic representations rather than an artifact of output 
formatting.

\section{Qwen2-VL-7B-Instruct: Extended Setup and Results}
\label{app:qwen}

We also tested Qwen2-VL-7B-Instruct, a recent 7B-parameter vision--language model released by Alibaba’s Qwen team~\citep{wang2024qwen2vlenhancingvisionlanguagemodels}. 
Experiments were run with HuggingFace Transformers (v4.56.1) on LRZ HPC using NVIDIA H100 GPUs (80 GB). 
The model follows a multimodal encoder–decoder design, pairing a ViT-based vision encoder with an instruction-tuned large language model.

To probe Stroop-style conflicts, we used the same synthetic dataset as in App.~\ref{appendix:dataset} (100 stimuli: 10 congruent, 90 incongruent). 
Two prompt variants were tested:

\begin{itemize}
    \item \textbf{Main:} ``You will see a single English word rendered in a colored ink. Ignore the written word and answer ONLY the ink color as one lowercase color name.''
    \item \textbf{Alternative:} ``What is the ink color of the text in the image? Answer with one lowercase color word only.''
\end{itemize}

Performance on congruent items was perfect under both prompts (100\% ink and word alignment). 
In incongruent cases, however, outcomes differed sharply: the longer instruction yielded only 31.1\% ink accuracy, while the shorter variant nearly doubled it to 60.0\%. 
This indicates that even advanced instruction-tuned VLMs inherit Stroop-style text dominance, but their behavior is sensitive to how the task is phrased.

\begin{table}[!htbp]
\centering
\begin{tabular}{|l|c|c|}
\hline
\textbf{Condition} & \textbf{Main prompt (\%)} & \textbf{Alt prompt (\%)} \\
\hline
Congruent (n=10)   & 100.0 & 100.0 \\
Incongruent (n=90) & 31.1  & 60.0 \\
\hline
\end{tabular}
\caption{Qwen2-VL-7B-Instruct Stroop results under two prompts. Prompt choice strongly influences incongruent ink accuracy.}
\label{tab:qwen_appendix}
\end{table}

Overall, Qwen2-VL shows somewhat higher sensitivity to ink color compared to CLIP and SigLIP-2, but still defaults to the written word when word and ink conflict. The gap between the two prompt conditions highlights the role of instruction design in shaping model responses.

\subsection*{Prompt-Variation Experiment}

Regarding whether Qwen2-VL’s Stroop behavior reflects a 
representational limitation or an instruction-following issue, we conducted an extended 
prompt-variation experiment. Building on the two prompts reported in the main appendix 
(Main vs.\ Alt), we constructed a broader set of ten minimally differing instructions:

\begin{enumerate}
    \item[\textbf{p1}] ``What is the ink color of the text in the image? Answer with one lowercase color word only.''
    \item[\textbf{p2}] ``Identify only the color of the ink used in the image. Respond with one lowercase color word.''
    \item[\textbf{p3}] ``Given the image, state the ink color of the printed word. Use a single lowercase color name.''
    \item[\textbf{p4}] ``Look at the picture and name the color of the ink, not the written word. Reply with a lowercase color.''
    \item[\textbf{p5}] ``Return only the ink color shown in the image. Do not describe anything else.''
    \item[\textbf{p6}] ``State the ink color of the letters. One lowercase color word only.''
    \item[\textbf{p7}] ``Ink color only — answer with one simple lowercase color word.''
    \item[\textbf{p8}] ``What is the color of the letters in this picture? Reply with one lowercase color name.''
    \item[\textbf{p9}] ``Read the image and give only the ink color as a lowercase color word.''
\end{enumerate}

Each prompt was evaluated on the 10 congruent and 90 incongruent Stroop items.  
Per-prompt ink accuracies are reported below:

\begin{table}[!htbp]
\centering
\begin{tabular}{|c|c|}
\hline
\textbf{Prompt} & \textbf{Ink Preference (\%)} \\
\hline
p1 & 58.0 \\
p2 & 88.0 \\
p3 & 48.0 \\
p4 & 45.0 \\
p5 & 83.0 \\
p6 & 37.0 \\
p7 & 24.0 \\
p8 & 97.0 \\
p9 & 31.0 \\
\hline
\end{tabular}
\caption{Ink accuracy of Qwen2-VL-7B-Instruct under nine alternative prompt formulations (p1--p9).}
\end{table}

The results reveal variability: ink accuracy ranges from as low as 24\% 
(up to 37–48\% for several neutral phrasings) to as high as 88–97\% for the clearest 
and most narrowly targeted instructions (prompts p2, p5, p8). This confirms that 
Qwen2-VL \emph{is capable} of attending to ink color, but only when the prompt is 
formulated with sufficient specificity and minimal linguistic ambiguity. Unlike CLIP, whose word-dominance is fixed by the geometry of its embedding space, Qwen2-VL exhibits an \emph{instruction-sensitive} form of 
text bias. The underlying representation still favors the written word by default, 
but careful prompting can partially redirect the model’s attention toward ink color.

\section{Latent Space Analyses}
\label{app:clip_latent}

To complement the behavioral findings, this appendix examines CLIP’s internal image representations. We ask whether the text-over-color preference observed in Sec.~\ref{sec:clip_siglip_behavioral} is also reflected \emph{in the geometry} of the embedding space. We analyze (i) unguided shape vs.\ color dominance, (ii) modality-specific contributions using Representational Dissimilarity Matrices (RDMs), and 
(iii) concept sensitivity via clustering.

\subsection*{Word vs.\ Ink Dominance in Embedding Space}

Here we compare image embeddings are only to bare color-word text embeddings (e.g., ``red'', ``blue'') without sentence framing. For each Stroop image we compute cosine similarity to two anchors—the written \emph{word} and the \emph{ink color}—and label the case as \textit{word-aligned} or \textit{ink-aligned} accordingly.

\begin{figure}[htbp]
    \centering
    \includegraphics[width=0.55\textwidth]{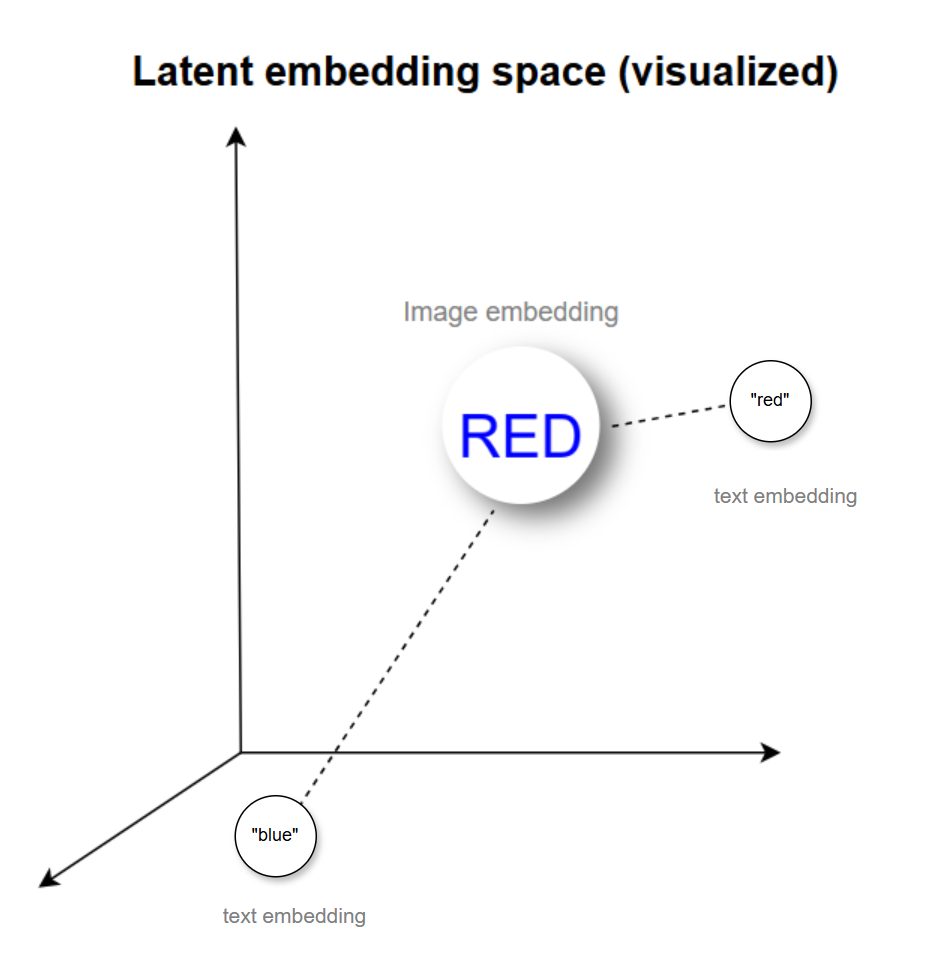}
    \caption{Unguided similarity test: the image embedding (e.g., ``RED'' written in blue) is compared to the concept anchors \textit{red} and \textit{blue} in text space; the closer anchor indicates the dominant modality.}
    \label{fig:embedding_space_color_vs_text}
\end{figure}

Across the 90 incongruent images, CLIP aligns with the \emph{written word} in 87 cases and with the \emph{ink color} in 3 (96.7\% vs.\ 3.3\%). Congruent images (10/10) align equally with both concepts as expected.


\noindent  Even without prompt steering, CLIP’s image embeddings sit closer to \emph{textual} concepts than to ink colors when the two conflict.

\subsection*{Modality Impact via Representational Dissimilarity Matrices}

We used Representational Dissimilarity Matrices (RDMs) to study how CLIP encodes \emph{word} and \emph{ink} information separately in the image embedding space.

We build RDMs (pairwise cosine dissimilarity) for three types of image sets: 
input images with only solid color \textbf{Ink-only}, images with only grayscale words \textbf{Word-only} 
(grayscale words), images that contains colored words \textbf{Word+Ink}.

The RDM for word-only means the representation similarity between images that just contain the word with identical gray color. The diagonal of the RDM is 0 which means the same word shape is the most similar to images with the same word shape in the embedding space. Similarly, the RDM for color only shows that the same colored image is the most similar to itself - while colors such as green elicits model representation that are very different from pure colored images such as black or gray or brown. 

To see how the shape of the word differentially contributes to the representational dissimilarity of the Stroop images, we subtract the ink-only RDM from the Stroop images. The delta RDM suggests how much the incremental information such as the shape of the word, shifts the RDM from the color to the colored word. Higher delta RDM means that there is a bigger change from the previous unimodal embedding disimilarity to the current bimodal embedding similarity. 

As shown in Figure~\ref{fig:featuresaliency}, when adding 
the shape of the word to the color, the RDM changes more compared to when shifting from having just the word shape to colored words. 

\[
\Delta\text{Word} = \text{RDM(Word+Ink)} - \text{RDM(Ink-only)}, 
\]
\[
\Delta\text{Ink} = \text{RDM(Word+Ink)} - \text{RDM(Word-only)}.
\]

\begin{figure}[htbp]
    \centering
    \includegraphics[width=0.60\textwidth]{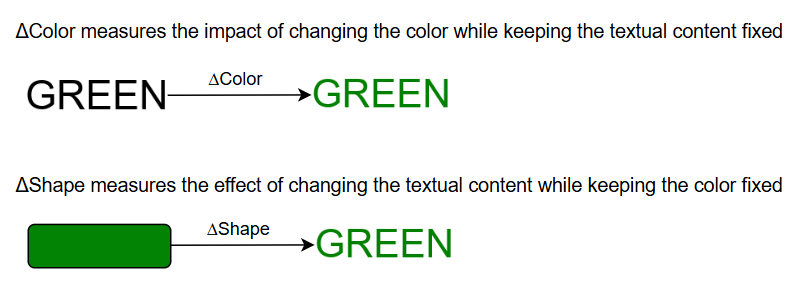}
    \caption{Design for modality-isolating RDM comparisons. $\Delta$Word/Shape asks ``what changes when text is added?'', $\Delta$Ink/Color asks ``what changes when color is added?''}
    \label{fig:delta_color_shape}
\end{figure}

\begin{figure}[htbp]
\centering
\begin{subfigure}[t]{0.45\columnwidth}
    \centering
    \includegraphics[height=4.2cm]{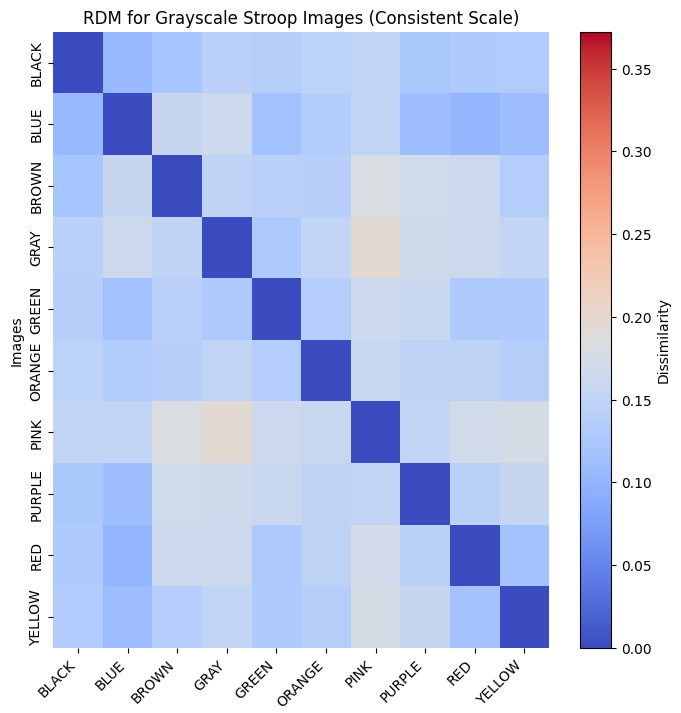}
    \caption{Word-only}
\end{subfigure}
\hfill
\begin{subfigure}[t]{0.45\columnwidth}
    \centering
    \includegraphics[height=4.2cm]{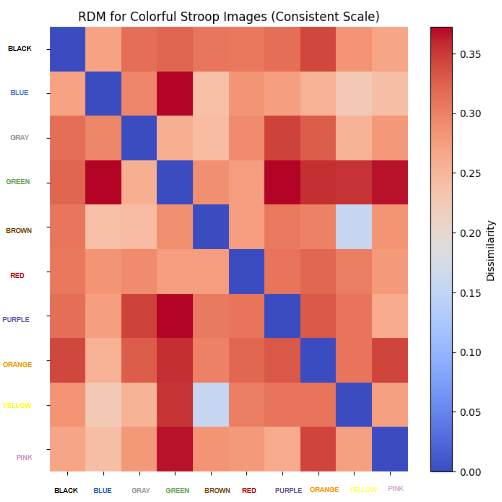}
    \caption{Word+Ink}
\end{subfigure}
\caption{RDMs for grayscale vs.\ colorful Stroop images.}
\label{fig:rdm_delta_color}
\end{figure}

\begin{figure}[htbp]
\centering
\begin{subfigure}[t]{0.45\columnwidth}
    \centering
    \includegraphics[height=4.2cm]{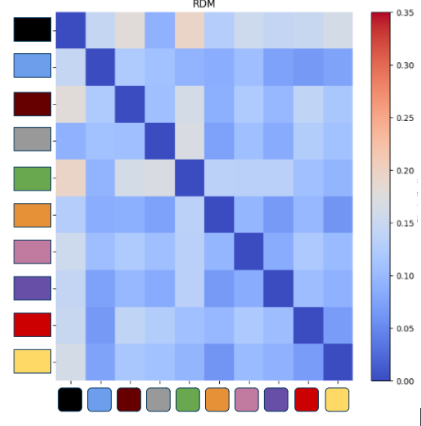}
    \caption{Ink-only}
\end{subfigure}
\hfill
\begin{subfigure}[t]{0.45\columnwidth}
    \centering
    \includegraphics[height=4.2cm]{images/rdm_colorful.png}
    \caption{Word+Ink}
\end{subfigure}
\caption{RDMs for solid-color vs.\ colorful text images.}
\label{fig:rdm_delta_shape}
\end{figure}


\noindent Adding \emph{word} reorganizes the space more sharply than adding \emph{ink color}, mirroring the behavioral text/shape preference.

\subsection*{Clustering and Dissimilarity Grouping}

We quantify per-color sensitivity to added text by the mean change in dissimilarity between Ink-only and Word+Ink embeddings, then cluster colors via agglomerative (Ward) linkage on cosine distance.

\begin{figure}[htbp]
\centering
\includegraphics[width=\columnwidth]{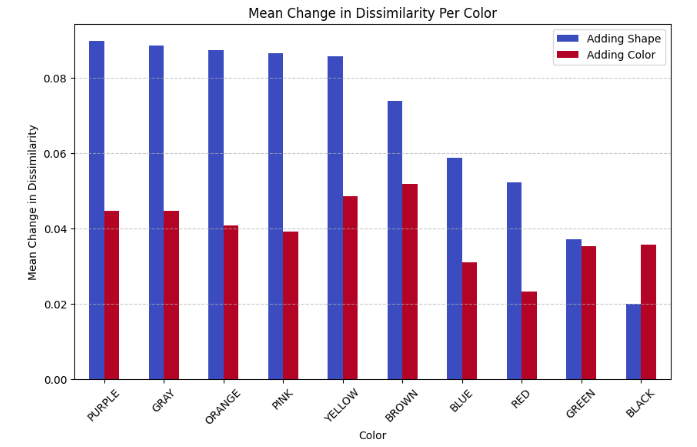}
\caption{Mean differential dissimilarity change when adding word form (shape) vs.\ color. Adding information about the word form in the image on average changes dissimilarity more strongly than adding the color to the words. }
\label{fig:mean_dissimilarity_per_color}
\end{figure}

Text dominance is \emph{systemic but not uniform}: CLIP’s reliance on shape varies with the color concept.

\subsection*{Representative CLIP Outputs}

Exemplar cosine-similarity outcomes under incongruent stimuli (winner among family-specific scores). 

\begin{table*}[H]
\centering
\begin{tabular}{|l|c|c|c|c|}
\hline
\textbf{Stimulus} & \textbf{Ground Truth (Ink)} & \textbf{Top-1 Text} & \textbf{Top-1 Color} & \textbf{Bias} \\
\hline
``RED'' in blue ink   & Blue   & Red (0.89)   & Blue (0.12)  & Textual \\
``GREEN'' in pink ink & Pink   & Green (0.76) & Pink (0.24)  & Mixed   \\
``BLACK'' in yellow ink & Yellow & Black (0.82) & Yellow (0.10) & Textual \\
\hline
\end{tabular}
\caption{Example CLIP similarity outputs on incongruent Stroop stimuli (cosine scores in parentheses).}
\label{tab:clip_outputs}
\end{table*}
Despite being asked for ink (Sec.~\ref{sec:clip_siglip_behavioral}), CLIP tends to output the textual label.


\section{Latent Intervention}
\label{app:latent_intervention}

This appendix provides extended results and implementation details for the latent intervention experiments. This include subpopulation-based chunk extraction, cosine-shift success, vector norms, and robustness to prompt variation.

\subsection*{Experimental Setup and Evaluation Protocol}
\label{app:latent_setup}

\textbf{Embeddings.} We use CLIP’s image encoder to obtain $d$–dimensional, $\ell_2$–normalized image embeddings. After any edit, we re-project to the unit sphere: $\mathbf{E}' \leftarrow \mathbf{E}'/\|\mathbf{E}'\|_2$ so that cosine similarity remains well-defined.

\textbf{Concept sets.} Chunks are built from \emph{congruent} images only (10 per class; App.~\ref{appendix:dataset}). For each source--target pair we intervene on 90 \emph{incongruent} sources $\times$ 9 targets.

Cosine similarity is computed to two prompt families (App.~\ref{appendix:dataset}): color–oriented and text–oriented. The shift metric is
\[
\Delta=\operatorname{sim}(\mathbf{E}',\text{target})-\operatorname{sim}(\mathbf{E}',\text{source}),
\]
and we count a \emph{success} if $\Delta>0$.

Unless noted, edits use scale $\alpha=1$. For completeness we also report robustness to $\alpha$ (see Sec.~\ref{app:alpha}).

\subsection{Subpopulation-Based Chunk Extraction}
\label{app:chunks}

To manipulate semantic concepts we derive directional vectors (\emph{chunks}) representing interpretable transitions (e.g., ``red'' $\rightarrow$ ``blue''). For each concept $c$, we compute per-dimension variance across its $N$ embeddings; cluster the variance values with $k$–means ($k=2$); and keep the \emph{low-variance} cluster as a binary mask $M_c \in \{0,1\}^d$. Stable means are
\[
\mu_c^{\text{stab}}=\frac{1}{N}\sum_i (M_c \odot \mathbf{E}_i^{(c)}),
\]
and the chunk vector is
\[
\vec{c}_{s\to t}^{\text{stab}} = \mu_t^{\text{stab}}-\mu_s^{\text{stab}}.
\]

We apply edits with optional scale $\alpha$:
\[
\mathbf{E}'=\operatorname{norm}\!\left(\mathbf{E}-\alpha\,\mu_s^{\text{stab}}+\alpha\,\mu_t^{\text{stab}}\right)
\]

Where $\operatorname{norm}(\cdot)$ reprojects to the unit sphere.

\begin{figure}[t]
\centering
\includegraphics[width=0.84\linewidth]{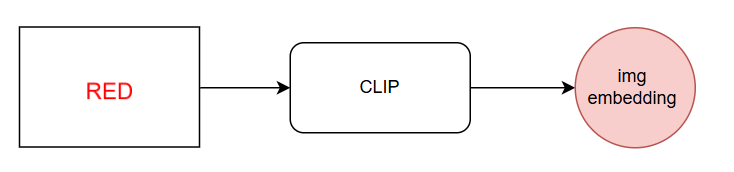}\\[0.6em]
\includegraphics[width=1.00\linewidth]{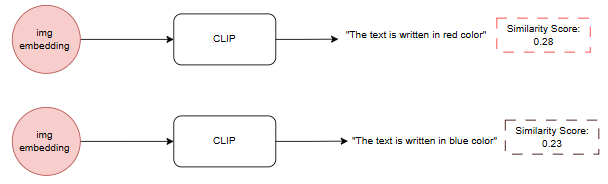}\\[0.6em]
\includegraphics[width=1.00\linewidth]{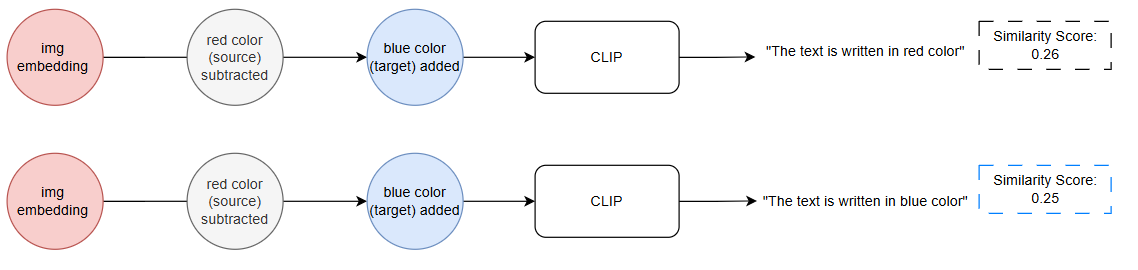}
\caption{Subpopulation-based intervention pipeline (RED$\to$BLUE). After editing we always re-normalize embeddings.}
\label{fig:latent_intervention_phases_app}
\end{figure}

\subsection{Subpopulation-Based Intervention Results}
\label{app:subpop_results}

\begin{table}[htbp]
\centering
\caption{Success and shift for subpopulation-based interventions}
\label{tab:D3_subpop}
\begin{tabular}{lcc}
\toprule
\textbf{Intervention} & \textbf{Success rate} & \textbf{Mean $\Delta$} \\
\midrule
Color    & 36.7\%  & $-0.0017$ \\
Text     & 100.0\% & $+0.0934$ \\
Combined & 100.0\% & $+0.1172$ \\
\bottomrule
\end{tabular}
\end{table}

\subsection{Chunk Vector Norm Strength}
\label{app:norms}

\begin{table}[H]
\centering
\caption{Chunk $\ell_2$ norms (color vs.\ text); mean across pairs shows $\sim$2.1$\times$ stronger text directions}
\label{tab:D4_chunk_norms}
\begin{tabular}{lccc}
\toprule
\textbf{Source$\to$Target} & \textbf{Color norm} & \textbf{Text norm} & \textbf{Notes} \\
\midrule
red$\to$blue      & 2.53 & 5.38 &  \\
red$\to$pink      & 3.60 & 6.38 &  \\
orange$\to$brown  & 2.09 & 6.61 &  (min color) \\
pink$\to$green    & 3.86 & 6.49 &  (max color) \\
\midrule
\textbf{Mean}     & \textbf{2.99} & \textbf{6.36} &  \\
\bottomrule
\end{tabular}
\end{table}

\subsection{Prompt Variations}
\label{app:prompts}

\begin{table}[H]
\centering
\caption{Effect of prompt variants on color-only intervention success rate}
\label{tab:D5_prompt_variants}
\begin{tabular}{llc}
\toprule
\textbf{Variant} & \textbf{Template} & \textbf{Success Rate} \\
\midrule
\texttt{color}     & The text is written in \{color\} color & 36.7\% \\
\texttt{just}      & \{color\}                              & 41.1\% \\
\texttt{pure}      & The text is purely \{color\}           & 41.1\% \\
\texttt{long}      & The color of the text is purely \{color\} & 38.9\% \\
\bottomrule
\end{tabular}
\end{table}

\subsection{Decomposing Combined Interventions}
\label{app:combined_decomp}
\begin{table}[H]
\centering
\caption{Contribution of color chunk within combined edits}
\label{tab:D6_color_contribution}
\begin{tabular}{lc}
\toprule
\textbf{Metric} & \textbf{Value} \\
\midrule
Mean $\Delta$ (Combined vs.\ Text-only) & $+0.0069$ \\
Cases with $|\Delta|<0.01$               & 60\% \\
Mean color-chunk norm                    & 3.04 \\
\bottomrule
\end{tabular}
\end{table}

\subsection{Robustness to Edit Scale $\alpha$}
\label{app:alpha}

We swept $\alpha\in\{0.25,0.5,1.0,1.5,2.0\}$. Text and combined edits improve monotonically up to $\alpha\approx1.5$ and then saturate; color-only edits remain near chance and become unstable for $\alpha>1.5$ (occasional regressions).

\subsection{Summary}
\label{app:summary}

\begin{itemize}
\item CLIP’s latent space is \textbf{modular for text}, but \textbf{entangled for color}. Text chunks are $\sim$2.1$\times$ stronger in norm and yield 100\% success; color-only remains weak even after subpopulation filtering.
\item Subpopulation filtering yields stable chunk vectors, but color interventions still succeed in only 36.7\% of cases with near-zero mean shifts.
\item Prompt variants provide small gains; scaling $\alpha$ mainly benefits text/combined edits.
\item Combined success is driven almost entirely by the text component.
\end{itemize}

\section{FLUX-Generated Realistic Multimodal Conflict Set}
\label{app:flux_examples}
This appendix reports the full set of multimodal conflict images generated using the FLUX.1 model, 
together with CLIP’s similarity scores under word- and color-oriented prompts. 
Unlike the controlled Stroop stimuli used in the main paper, these examples capture 
realistic forms of semantic contradiction that arise in everyday visual interfaces 
(e.g., signage, UI icons, safety symbols, battery indicators, or traffic signals). 
Each image was paired with two prompts—the written word (semantic cue) and the dominant color 
(visual cue)—allowing us to directly test whether CLIP prioritizes textual or visual information 
when the two conflict.

Overall, CLIP exhibits a mixed pattern: text-heavy or OCR-like stimuli tend to trigger strong 
word dominance, whereas chromatically salient or icon-based stimuli shift the model toward color-based decisions. 

Table~\ref{tab:flux_examples_full} summarizes ten representative cases from our evaluation set, 
showing both the raw similarity scores and the resulting decision for each conflict instance. 
These examples illustrate that the Stroop-style word bias documented in the synthetic setting 
generalizes only partially: in realistic conflict scenarios, visual salience and icon semantics 
can override the written word. 

\begin{table*}[t]
\centering
\caption{\textbf{FLUX-generated multimodal conflict examples.} 
CLIP’s decision reflects either the textual or visual cue depending on saliency.}
\label{tab:flux_examples_full}

\setlength{\tabcolsep}{4pt}
\begin{tabular}{c c c c c}
\toprule
\textbf{Image} & \textbf{Visual Description} & \textbf{word\_sim} & \textbf{color\_sim} & \textbf{Decision} \\
\midrule

\raisebox{-0.5\height}{\includegraphics[width=0.12\textwidth]{images/samples_1.png}}
& STOP (green background) & 0.17 & 0.83 & COLOR \\[1mm]
\midrule

\raisebox{-0.5\height}{\includegraphics[width=0.12\textwidth]{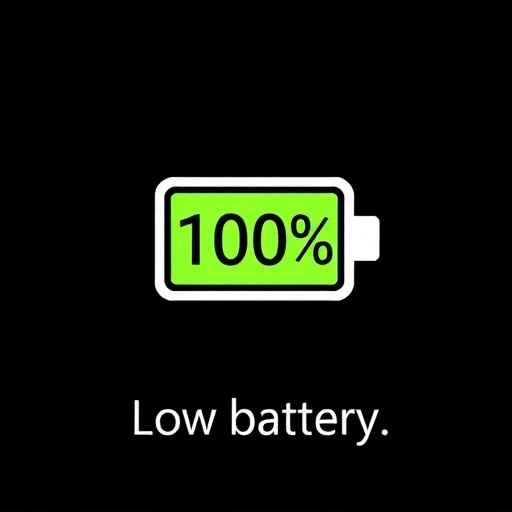}}
& Low Battery (green icon) & 0.03 & 0.96 & COLOR \\[1mm]
\midrule

\raisebox{-0.5\height}{\includegraphics[width=0.12\textwidth]{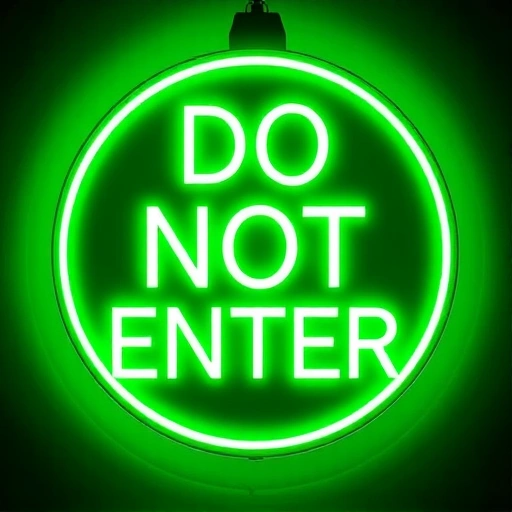}}
& DO NOT ENTER (green sign) & 0.9469 & 0.0530 & WORD \\[1mm]
\midrule

\raisebox{-0.5\height}{\includegraphics[width=0.12\textwidth]{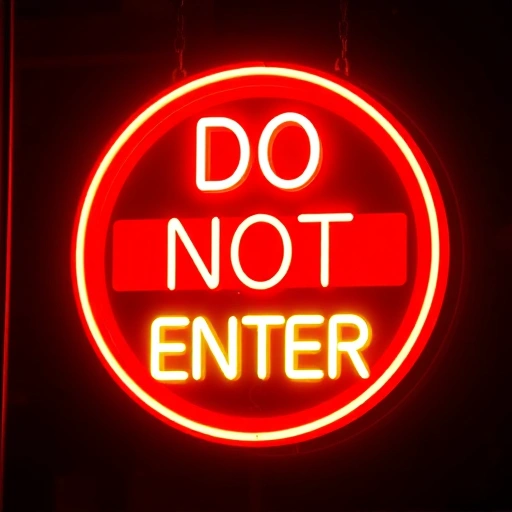}}
& DO NOT ENTER (red sign) & 0.9953 & 0.0047 & WORD \\[1mm]
\midrule

\raisebox{-0.5\height}{\includegraphics[width=0.12\textwidth]{images/samples_8.png}}
& EXIT (red sign) & 0.9988 & 0.0011 & WORD \\[1mm]
\midrule

\raisebox{-0.5\height}{\includegraphics[width=0.12\textwidth]{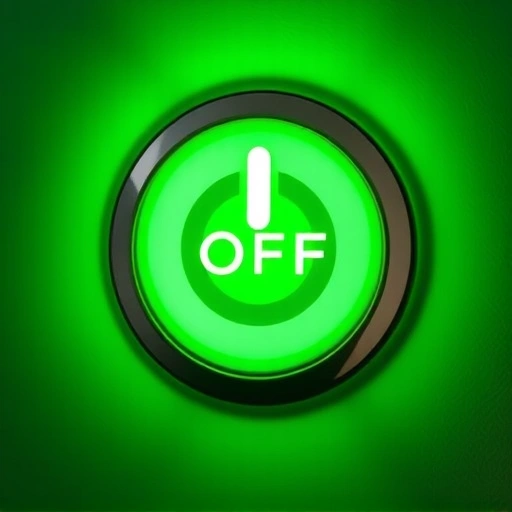}}
& OFF (green neon) & 0.0020 & 0.9979 & COLOR \\[1mm]
\midrule

\raisebox{-0.5\height}{\includegraphics[width=0.12\textwidth]{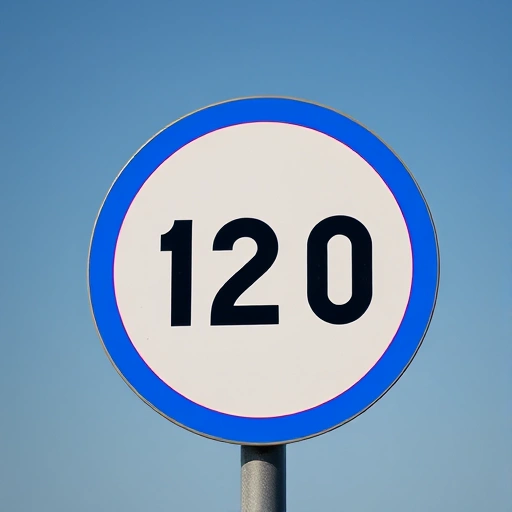}}
& 120 speed limit (blue) & 0.9992 & 0.0007 & WORD \\[1mm]
\midrule

\raisebox{-0.5\height}{\includegraphics[width=0.12\textwidth]{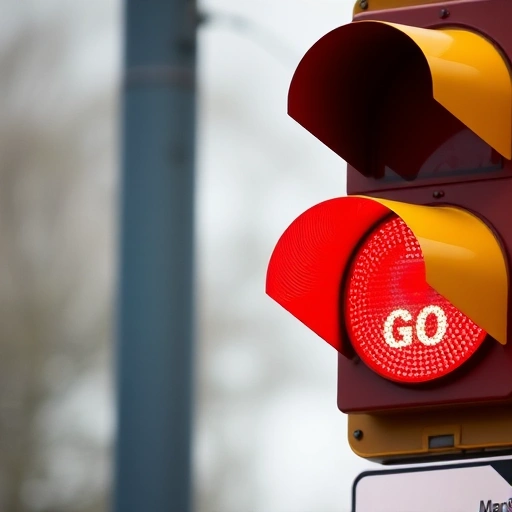}}
& GO (red traffic light) & 0.2223 & 0.7777 & COLOR \\[1mm]
\midrule

\raisebox{-0.5\height}{\includegraphics[width=0.12\textwidth]{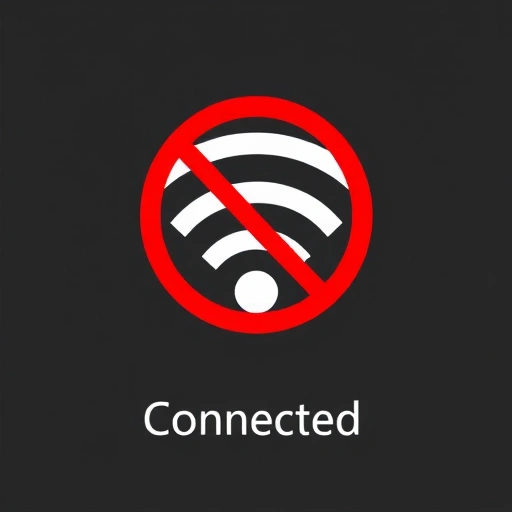}}
& CONNECTED (red crossed-out wifi) & 0.00015 & 0.9998 & COLOR \\[1mm]
\midrule

\raisebox{-0.5\height}{\includegraphics[width=0.12\textwidth]{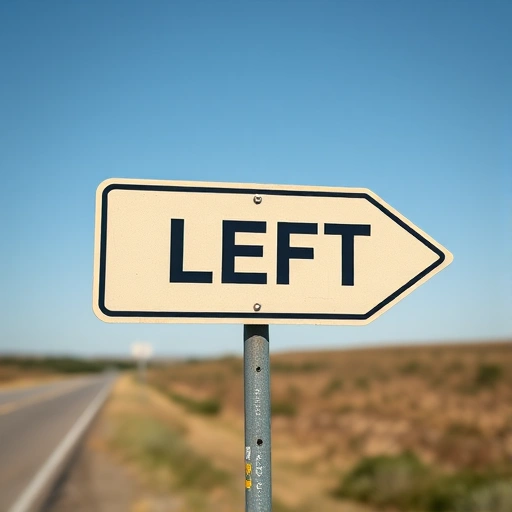}}
& LEFT (arrow pointing right) & 0.9445 & 0.0554 & WORD \\
\bottomrule
\end{tabular}
\end{table*}



\end{document}

%% file: math_commands.tex
\usepackage{amsmath,amsfonts,bm}

\def\eqref#1{equation~\ref{#1}}

\def\1{\bm{1}}

\DeclareMathAlphabet{\mathsfit}{\encodingdefault}{\sfdefault}{m}{sl}
\SetMathAlphabet{\mathsfit}{bold}{\encodingdefault}{\sfdefault}{bx}{n}



%% file: aaai2027.bib
@inproceedings{xiao2025flair,
  title={FLAIR: VLM with Fine-grained Language-informed Image Representations},
  author={Xiao, Rui and Kim, Sanghwan and Georgescu, Mariana-Iuliana and Akata, Zeynep and Alaniz, Stephan},
  booktitle={CVPR},
  year={2025}
}

@inproceedings{kim2025cosmos,
  title={COSMOS: Cross-Modality Self-Distillation for Vision Language Pre-training},
  author={Kim, Sanghwan and Xiao, Rui and Georgescu, Mariana-Iuliana and Alaniz, Stephan and Akata, Zeynep},
  booktitle={CVPR},
  year={2025}
}

@inproceedings{li2023pope,
  title={Evaluating object hallucination in large vision-language models},
  author={Li, Yifan and Du, Yifan and Zhou, Kun and Wang, Jinpeng and Zhao, Xin and Wen, Ji-Rong},
  booktitle={EMNLP},
  year={2023}
}

@article{zhao2025wcit,
  title   = {What Color Is the Text? {A} Benchmark for Hallucination 
             Induced by Image-Embedded Prompt},
  author  = {Zhao, Jinkun and Huang, Lei and Ge, Haixin and Wu, Wenjun},
  journal = {arXiv preprint arXiv:2511.13400},
  year    = {2025}
}

@misc{deng2025wordsvisionvisionlanguagemodels,
      title={Words or Vision: Do Vision-Language Models Have Blind Faith in Text?}, 
      author={Ailin Deng and Tri Cao and Zhirui Chen and Bryan Hooi},
      year={2025},
      eprint={2503.02199},
      archivePrefix={arXiv},
      primaryClass={cs.CV},
      url={https://arxiv.org/abs/2503.02199}, 
}

@article{navon1977353,
title = {Forest before trees: The precedence of global features in visual perception},
journal = {Cognitive Psychology},
volume = {9},
number = {3},
pages = {353-383},
year = {1977},
issn = {0010-0285},
doi = {https://doi.org/10.1016/0010-0285(77)90012-3},
url = {https://www.sciencedirect.com/science/article/pii/0010028577900123},
author = {David Navon}
}

@misc{Radford2021,
  title         = {Learning Transferable Visual Models From Natural Language Supervision},
  author        = {Alec Radford and Jong Wook Kim and Chris Hallacy and Aditya Ramesh and Gabriel Goh and Sandhini Agarwal and Girish Sastry and Amanda Askell and Pamela Mishkin and Jack Clark and Gretchen Krueger and Ilya Sutskever},
  year          = {2021},
  archivePrefix = {arXiv},
  eprint        = {2103.00020},
  primaryClass  = {cs.CV},
  url           = {https://arxiv.org/abs/2103.00020}
}

@inproceedings{Xu2015,
  title     = {Show, Attend and Tell: Neural Image Caption Generation with Visual Attention},
  author    = {Xu, Kelvin and Ba, Jimmy and Kiros, Ryan and Cho, Kyunghyun and Courville, Aaron and Salakhudinov, Ruslan and Zemel, Richard and Bengio, Yoshua},
  booktitle = {Proceedings of the 32nd International Conference on Machine Learning},
  series    = {Proceedings of Machine Learning Research},
  volume    = {37},
  pages     = {2048--2057},
  year      = {2015},
  organization = {PMLR},
  url       = {https://proceedings.mlr.press/v37/xuc15.html}
}

@article{Hossain2019,
  title     = {A Comprehensive Survey of Deep Learning for Image Captioning},
  author    = {Hossain, Md Zia Uddin and Sohel, Ferdous and Shiratuddin, Mohd Fairuz and Laga, Hamid},
  journal   = {ACM Computing Surveys (CSUR)},
  volume    = {51},
  number    = {6},
  pages     = {1--36},
  year      = {2019},
  publisher = {ACM},
  doi       = {10.1145/3295748},
  url       = {https://dl.acm.org/doi/10.1145/3295748}
}

@article{Brascamp2021,
   author = "Brascamp, Jan W. and Shevell, Steven K.",
   title = "The Certainty of Ambiguity in Visual Neural Representations", 
   journal= "Annual Review of Vision Science",
   year = "2021",
   volume = "7",
   number = "Volume 7, 2021",
   pages = "465-486",
   doi = "https://doi.org/10.1146/annurev-vision-100419-125929",
   url = "https://www.annualreviews.org/content/journals/10.1146/annurev-vision-100419-125929",
   publisher = "Annual Reviews",
   issn = "2374-4650",
   type = "Journal Article",
  }

@article{Tanaka1999,
  author  = {Tanaka, James W. and Presnell, Lynn M.},
  title   = {Color diagnosticity in object recognition},
  journal = {Perception \& Psychophysics},
  volume  = {61},
  number  = {6},
  pages   = {1140--1153},
  year    = {1999},
  month   = aug,
  issn    = {1532-5962},
  doi     = {10.3758/BF03207619},
  url     = {https://doi.org/10.3758/BF03207619}
}

@article{landau1988,
title = {The importance of shape in early lexical learning},
journal = {Cognitive Development},
volume = {3},
number = {3},
pages = {299-321},
year = {1988},
issn = {0885-2014},
doi = {https://doi.org/10.1016/0885-2014(88)90014-7},
url = {https://www.sciencedirect.com/science/article/pii/0885201488900147},
author = {Barbara Landau and Linda B. Smith and Susan S. Jones}
}

@incollection{smith2000,
    author = {Smith, Linda B.},
    editor = {Golinkoff, Roberta Micknick and Hirsh-Pasek, Kathryn},
    isbn = {9780195130324},
    title = {Learning How to Learn Words: An Associative Crane},
    booktitle = {Becoming a Word Learner: A Debate on Lexical Acquisition},
    publisher = {Oxford University Press},
    year = {2000},
    month = {12},
    doi = {10.1093/acprof:oso/9780195130324.003.003},
    url = {https://doi.org/10.1093/acprof:oso/9780195130324.003.003},
    eprint = {https://academic.oup.com/book/0/chapter/159226169/chapter-ag-pdf/44932813/book_10946_section_159226169.ag.pdf},
}

@ARTICLE{Kornmeier2012,
    
AUTHOR={Kornmeier, Jürgen  and Bach, Michael },
           
TITLE={Ambiguous Figures – What Happens in the Brain When Perception Changes But Not the Stimulus},
          
JOURNAL={Frontiers in Human Neuroscience},
        
volume = {6},
  
YEAR={2012},
  
URL={https://www.frontiersin.org/journals/human-neuroscience/articles/10.3389/fnhum.2012.00051},
  
DOI={10.3389/fnhum.2012.00051},
  
ISSN={1662-5161}}

@inproceedings{Faghri2018,
  title        = {VSE++: Improving Visual-Semantic Embeddings with Hard Negatives},
  author       = {Faghri, Fartash and Fleet, David J. and Kiros, Jamie Ryan and Fidler, Sanja},
  booktitle    = {Proceedings of the British Machine Vision Conference (BMVC)},
  year         = {2018},
  organization = {BMVA},
  url          = {https://bmva-archive.org.uk/bmvc/2018/contents/papers/0344.pdf},
  note         = {BMVC Spotlight paper}
}

@inproceedings{Antol2015,
  title        = {VQA: Visual Question Answering},
  author       = {Antol, Stanislaw and Agrawal, Aishwarya and Lu, Jiasen and Mitchell, Margaret and Batra, Dhruv and Zitnick, C. Lawrence and Parikh, Devi},
  booktitle    = {Proceedings of the IEEE International Conference on Computer Vision (ICCV)},
  month        = {December},
  year         = {2015},
  organization = {IEEE/CVF},
  doi          = {10.1109/ICCV.2015.279},
}

@inproceedings{Alayrac2022,
  title     = {Flamingo: A Visual Language Model for Few-Shot Learning},
  author    = {Alayrac, Jean-Baptiste and Donahue, Jeff and Luc, Pauline and Miech, Antoine and Barr, Iain and Hasson, Yana and Lenc, Karel and Mensch, Arthur and Millican, Katie and Reynolds, Malcolm and Ring, Roman and Rutherford, Eliza and Cabi, Serkan and Han, Tengda and Gong, Zhitao and Samangooei, Sina and Monteiro, Marianne and Menick, Jacob L. and Borgeaud, Sebastian and Brock, Andrew and Nematzadeh, Aida and Sharifzadeh, Sahand and Binkowski, Mikolaj and Barreira, Ricardo and Vinyals, Oriol and Zisserman, Andrew and Simonyan, Karen},
  booktitle = {Advances in Neural Information Processing Systems},
  series    = {NeurIPS ’22},
  year      = {2022},
  url       = {https://proceedings.neurips.cc/paper_files/paper/2022/hash/960a172bc7fbf0177ccccbb411a7d800-Abstract-Conference.html}
}

@inproceedings{Tsimpoukelli2021,
  title        = {Multimodal Few-Shot Learning with Frozen Language Models},
  author       = {Tsimpoukelli, Maria and Menick, Jacob and Cabi, Serkan and Eslami, S. M. Ali and Vinyals, Oriol and Hill, Felix},
  booktitle    = {Advances in Neural Information Processing Systems},
  series       = {NeurIPS},
  volume       = {34},
  pages        = {200--212},
  year         = {2021},
  url          = {https://proceedings.neurips.cc/paper/2021/hash/01b7575c38dac42f3cfb7d500438b875-Abstract.html},
  arxiv        = {2106.13884},
  doi          = {10.48550/arXiv.2106.13884}
}

@misc{Pach2025,
  title        = {Sparse Autoencoders Learn Monosemantic Features in Vision-Language Models},
  author       = {Mateusz Pach and Shyamgopal Karthik and Quentin Bouniot and Serge Belongie and Zeynep Akata},
  year         = {2025},
  archivePrefix= {arXiv},
  eprint       = {2504.02821},
  primaryClass = {cs.CV},
  url          = {https://arxiv.org/abs/2504.02821}
}

@article{Stroop1935,
  title={Studies of Interference in Serial Verbal Reactions},
  author={Stroop, John Ridley},
  journal={Journal of Experimental Psychology},
  volume={18},
  number={6},
  pages={643--662},
  year={1935},
  doi={10.1037/h0054651}
}

@article{Arias_2024,
   title={Color in Visual-Language Models: CLIP Deficiencies},
   volume={32},
   ISSN={2166-9635},
   url={http://dx.doi.org/10.2352/CIC.2024.32.1.20},
   DOI={10.2352/cic.2024.32.1.20},
   number={1},
   journal={Color and Imaging Conference},
   publisher={Society for Imaging Science & Technology},
   author={Arias, Guillem and Baldrich, Ramon and Vanrell, Maria},
   year={2024},
   month=oct, pages={101–106} }

@inproceedings{
dai2023instructblip,
title={Instruct{BLIP}: Towards General-purpose Vision-Language Models with Instruction Tuning},
author={Wenliang Dai and Junnan Li and Dongxu Li and Anthony Tiong and Junqi Zhao and Weisheng Wang and Boyang Li and Pascale Fung and Steven Hoi},
booktitle={Thirty-seventh Conference on Neural Information Processing Systems},
year={2023},
url={https://openreview.net/forum?id=vvoWPYqZJA}
}

@inproceedings{liu2023visualinstructiontuning,
 author    = {Liu, Haotian and Li, Chunyuan and Wu, Qingyang and Lee, Yong Jae},
 booktitle = {Advances in Neural Information Processing Systems},
 editor    = {A. Oh and T. Naumann and A. Globerson and K. Saenko and M. Hardt and S. Levine},
 pages     = {34892--34916},
 publisher = {Curran Associates, Inc.},
 title     = {Visual Instruction Tuning},
 url       = {https://proceedings.neurips.cc/paper_files/paper/2023/file/6dcf277ea32ce3288914faf369fe6de0-Paper-Conference.pdf},
 volume    = {36},
 year      = {2023}
}

@article{Wang2022GIT,
  title   = {{GIT}: A Generative Image-to-text Transformer for Vision and Language},
  author  = {Jianfeng Wang and Zhengyuan Yang and Xiaowei Hu and Linjie Li and 
             Kevin Lin and Zhe Gan and Zicheng Liu and Ce Liu and Lijuan Wang},
  journal = {Transactions on Machine Learning Research},
  issn    = {2835-8856},
  year    = {2022},
  url     = {https://openreview.net/forum?id=b4tMhpN0JC}
}

@inproceedings{wu2025discoveringchunksneuralembeddings,
  title     = {Concept-Guided Interpretability via Neural Chunking},
  author    = {Shuchen Wu and Stephan Alaniz and Shyamgopal Karthik and 
               Peter Dayan and Eric Schulz and Zeynep Akata},
  booktitle = {The Thirty-ninth Annual Conference on Neural Information Processing Systems},
  year      = {2025},
  url       = {https://openreview.net/forum?id=o87dDXYLXC}
}

@inproceedings{
peng2024grounding,
title={Grounding Multimodal Large Language Models to the World},
author={Zhiliang Peng and Wenhui Wang and Li Dong and Yaru Hao and Shaohan Huang and Shuming Ma and Qixiang Ye and Furu Wei},
booktitle={The Twelfth International Conference on Learning Representations},
year={2024},
url={https://openreview.net/forum?id=lLmqxkfSIw}
}

@inproceedings{Li2022BLIP,
  title     = {{BLIP}: Bootstrapping Language-Image Pre-training for 
               Unified Vision-Language Understanding and Generation},
  author    = {Li, Junnan and Li, Dongxu and Xiong, Caiming and Hoi, Steven},
  booktitle = {Proceedings of the 39th International Conference on Machine Learning},
  pages     = {12888--12900},
  year      = {2022},
  editor    = {Chaudhuri, Kamalika and Jegelka, Stefanie and Song, Le and 
               Szepesvari, Csaba and Niu, Gang and Sabato, Sivan},
  volume    = {162},
  series    = {Proceedings of Machine Learning Research},
  month     = {17--23 Jul},
  publisher = {PMLR},
  url       = {https://proceedings.mlr.press/v162/li22n.html}
}

@article{Kriegeskorte2008,
  author    = {Kriegeskorte, Nikolaus and Mur, Marieke and Bandettini, Peter},
  title     = {Representational similarity analysis – connecting the branches of systems neuroscience},
  journal   = {Frontiers in Systems Neuroscience},
  volume    = {2},
  pages     = {4},
  year      = {2008},
  doi       = {10.3389/neuro.06.004.2008},
  url       = {https://www.frontiersin.org/articles/10.3389/neuro.06.004.2008/full}
}

@inproceedings{
  yuksekgonul2023when,
  title={When and why Vision-Language Models behave like  Bags-of-Words, and what to do about it?},
  author={Mert Yuksekgonul and Federico Bianchi and Pratyusha   Kalluri and Dan Jurafsky and James Zou},
  booktitle={International Conference on Learning Representations},
  year={2023},
  url={https://openreview.net/forum?id=KRLUvxh8uaX}
}

@article{Goh2021,
  title     = {Multimodal Neurons in Artificial Neural Networks},
  author    = {Goh, Gabriel and Petrov, Nick and Mayo, Chelsea and Regev, Michael and Carter, Shan and Olah, Christopher},
  journal   = {Distill},
  year      = {2021},
  doi       = {10.23915/distill.00030},
  url       = {https://distill.pub/2021/multimodal-neurons/},
  note      = {https://distill.pub/2021/multimodal-neurons/},
  publisher = {Distill}
}

@misc{tschannen2025siglip2multilingualvisionlanguage,
  title        = {SigLIP 2: Multilingual Vision-Language Encoders with Improved Semantic Understanding, Localization, and Dense Features},
  author       = {Michael Tschannen and Alexey Gritsenko and Xiao Wang and Muhammad Ferjad Naeem and Ibrahim Alabdulmohsin and Nikhil Parthasarathy and Talfan Evans and Lucas Beyer and Ye Xia and Basil Mustafa and Olivier Hénaff and Jeremiah Harmsen and Andreas Steiner and Xiaohua Zhai},
  year         = {2025},
  eprint       = {2502.14786},
  archivePrefix= {arXiv},
  primaryClass = {cs.CV},
  url          = {https://arxiv.org/abs/2502.14786},
}

@misc{wang2024qwen2vlenhancingvisionlanguagemodels,
  title={Qwen2-VL: Enhancing Vision-Language Model's Perception of the World at Any Resolution}, 
  author={Peng Wang and Shuai Bai and Sinan Tan and Shijie Wang and Zhihao Fan and Jinze Bai and Keqin Chen and Xuejing Liu and Jialin Wang and Wenbin Ge and Yang Fan and Kai Dang and Mengfei Du and Xuancheng Ren and Rui Men and Dayiheng Liu and Chang Zhou and Jingren Zhou and Junyang Lin},
  year={2024},
  eprint={2409.12191},
  archivePrefix={arXiv},
  primaryClass={cs.CV},
  url={https://arxiv.org/abs/2409.12191}
}

@misc{pezeshkpour2025mixedsignalsdecodingvlms,
      title={Mixed Signals: Decoding VLMs' Reasoning and Underlying Bias in Vision-Language Conflict}, 
      author={Pouya Pezeshkpour and Moin Aminnaseri and Estevam Hruschka},
      year={2025},
      eprint={2504.08974},
      archivePrefix={arXiv},
      primaryClass={cs.AI},
      url={https://arxiv.org/abs/2504.08974}, 
}

@inproceedings{zhai2023sigmoid,
  title={Sigmoid loss for language image pre-training},
  author={Zhai, Xiaohua and Mustafa, Basil and Kolesnikov, Alexander and Beyer, Lucas},
  booktitle={Proceedings of the IEEE/CVF international conference on computer vision},
  pages={11975--11986},
  year={2023}
}

@inproceedings{
geirhos2018imagenettrained,
title={ImageNet-trained {CNN}s are biased towards texture; increasing shape bias improves accuracy and robustness.},
author={Robert Geirhos and Patricia Rubisch and Claudio Michaelis and Matthias Bethge and Felix A. Wichmann and Wieland Brendel},
booktitle={International Conference on Learning Representations},
year={2019},
url={https://openreview.net/forum?id=Bygh9j09KX},
}

@misc{jia2025benchmarkingmultimodalknowledgeconflict,
      title={Benchmarking Multimodal Knowledge Conflict for Large Multimodal Models}, 
      author={Yifan Jia and Kailin Jiang and Yuyang Liang and Qihan Ren and Yi Xin and Rui Yang and Fenze Feng and Mingcai Chen and Hengyang Lu and Haozhe Wang and Xiaoye Qu and Dongrui Liu and Lizhen Cui and Yuntao Du},
      year={2025},
      eprint={2505.19509},
      archivePrefix={arXiv},
      primaryClass={cs.LG},
      url={https://arxiv.org/abs/2505.19509}, 
}

@misc{menon2022taskbiasvisionlanguagemodels,
      title={Task Bias in Vision-Language Models}, 
      author={Sachit Menon and Ishaan Preetam Chandratreya and Carl Vondrick},
      year={2022},
      eprint={2212.04412},
      archivePrefix={arXiv},
      primaryClass={cs.CV},
      url={https://arxiv.org/abs/2212.04412}, 
}

@inproceedings{deng2025wordsorvision,
  title={Words or Vision: Do Vision-Language Models Have Blind Faith in Text?},
  author={Deng, Ailin and Cao, Tri and Chen, Zhirui and Hooi, Bryan},
  booktitle={CVPR},
  year={2025}
}

@inproceedings{xiao2026finer,
  title={FINER: MLLMs Hallucinate under Fine-grained Negative Queries},
  author={Xiao, Rui and Kim, Sanghwan and Xian, Yongqin and Akata, Zeynep and Alaniz, Stephan},
  booktitle={CVPR},
  year={2026}
}
